\documentclass[conference,compsoc]{IEEEtran}
\usepackage{xcolor}
\usepackage{xspace}
\usepackage{comment}
\usepackage{amsmath}
\usepackage{subfigure}
\usepackage{graphicx}
\usepackage{multirow}
\usepackage[hidelinks]{hyperref}
\usepackage{array}
\usepackage{setspace}
\usepackage[hang,flushmargin]{footmisc} 

\ifCLASSOPTIONcompsoc
  \usepackage[nocompress]{cite}
\else
  \usepackage{cite}
\fi
\ifCLASSINFOpdf
\else
\fi
\begin{document}
\newcommand{\eg}{{\em e.g.},\xspace}
\newcommand{\ie}{{\em i.e.},\xspace}
\newcommand{\etal}{{\em et al.}\xspace}
\newcommand{\name}{\textsc{Loki}\xspace}
\newcommand{\namemeaning}{In Norse mythology, \name is a cunning trickster who had the ability to change his shape. Our attack can figuratively change its shape for different clients and trick them into giving up their private data.}
\newcommand{\atul}[1]{\textcolor{purple}{Atul: #1}}
\newcommand{\saurabh}[1]{\textcolor{red}{Saurabh: #1}}
\newcommand{\salman}[1]{\textcolor{blue}{Salman: #1}}

\newcommand{\chop}[1]{}
%
\title{\name: Large-scale Data Reconstruction Attack against Federated Learning through Model Manipulation}

%

\author{Joshua C. Zhao$^1$, Atul Sharma$^1$, Ahmed Roushdy Elkordy$^2$, Yahya H. Ezzeldin$^2$ \\Salman Avestimehr$^2$, Saurabh Bagchi$^1$\\
\small{$^1$Purdue University, $^2$University of Southern California}\\
{\tt\small \{zhao1207,sharm438,sbagchi\}@purdue.edu}, 
{\tt\small \{aelkordy,yessa,avestime\}@usc.edu}}



\maketitle

\begin{abstract}
Federated learning was introduced to enable machine learning over large decentralized datasets while promising privacy by eliminating the need for data sharing. Despite this, prior work has shown that shared gradients often contain private information and attackers can gain knowledge either through malicious modification of the architecture and parameters or by using optimization to approximate user data from the shared gradients. 

However, prior data reconstruction attacks have been limited in setting and scale, as most works target FedSGD and limit the attack to single-client gradients. Many of these attacks fail in the more practical setting of FedAVG or if updates are aggregated together using secure aggregation. Data reconstruction becomes significantly more difficult, resulting in limited attack scale and/or decreased reconstruction quality. When both FedAVG and secure aggregation are used, there is no current method that is able to attack multiple clients concurrently in a federated learning setting.

In this work we introduce \name, an attack that overcomes previous limitations and also breaks the anonymity of aggregation as the leaked data is identifiable and directly tied back to the clients they come from. Our design sends clients customized convolutional parameters,  and the weight gradients of data points between clients remain separate even through aggregation. 
With FedAVG and aggregation across 100 clients, prior work can leak less than 1\% of images on MNIST, CIFAR-100, and Tiny ImageNet.
Using only a single training round, \name is able to leak 76-86\% of all data samples.
\end{abstract}



%
\IEEEpeerreviewmaketitle

\section{Introduction}
\vspace{-2mm}
Federated learning (FL)~\cite{mcmahan2017communication} is a machine learning paradigm introduced to address growing user data privacy concerns where clients participate in large-scale model training without sending their private data to servers for centralized training. A general training round in FL consists of a server sending the model out to the clients and clients training the received model with their private data before sending their updates back to the server. The server then aggregates the updates from the clients and uses the aggregate to update the global model before sending it out to repeat another iteration of the training process. Two primary methods of training are FedSGD and FedAVG. FedSGD involves a client training a model for a single local iteration before sending the gradients to the server. FedAVG, on the other hand, involves several local iterations of client training before sending the updated model parameters to the server. Due to communication efficiency, FedAVG is preferred in many settings.

While the promise of FL is that user data should be private and secure from a malicious server, this only holds true if gradients from the clients cannot be used by a malicious server to infer properties of the training data. Prior work has shown that property inference~\cite{melis2019exploiting,luo2021feature}, membership inference~\cite{shokri2017membership,choquette2021label,nasr2019comprehensive} or GAN-based attacks~\cite{hitaj2017deep,wang2019beyond} are all effective in inferring information about client data from their shared gradients. However, the stronger class of data reconstruction attacks, when a malicious server wants to directly steal private client training data, has been demonstrated against even the strictest defenses. 
These attacks fall into two categories.

Optimization attacks~\cite{zhu2019deep,zhao2020idlg,geiping2020inverting,yin2021see} generally work on image data, starting with a dummy randomly initialized data sample and optimizing over the difference between the true gradient and the one generated through the dummy sample. Iteratively, the dummy data sample  is updated and gets closer to the ground truth data that was used in computing the true gradient. However reconstructions following this method degrade in quality as the batch size and image resolution increase, failing with a batch size of greater than 48 as demonstrated in~\cite{yin2021see} on the ImageNet dataset. Most optimization attacks work only on FedSGD, but~\cite{dimitrov2022data} demonstrates an optimization attack on FedAVG that is applicable to single client attacks with a relatively small client dataset size and image size (50 images in total, FEMNIST/CIFAR-100). This is promising, as FedAVG is inherently more difficult to attack since the malicious server cannot see the intermediate model updates coming from local iterations. 

However, secure aggregation has been shown to be an effective defense against all optimization-based privacy attacks in FL. 
Secure aggregation  guarantees that the server 
will not gain access to individual model updates of other clients, only the aggregate of all client updates~\cite{bonawitz2017practical,fereidooni2021safelearn,so2021lightsecagg}. This poses a large problem for optimization as aggregation introduces a massive number of total images in training while these attacks only have a high image reconstruction rate in the small batch size regime.

The second class of attacks, analytic reconstruction attacks~\cite{fowl2022robbing,boenisch2021curious}, involve customizing the model parameters or the model architecture to directly retrieve the training data from the gradients of a fully-connected (FC) layer --- this is referred to as {\em linear layer leakage}. These approaches do not have issues with quality as they reconstruct the inputs near exactly. These methods work well when individual client updates are visible. 
Under secure aggregation and FedSGD, ~\cite{fowl2022robbing} also has the ability to maintain a high leakage rate by increasing the size of the injected FC layer, although this often results in very large models. The introduced sparse variant of their attack functions to attack individual updates in FedAVG, but fails when secure aggregation is applied.

Some recent works have targeted FL with secure aggregation by magnifying gradients~\cite{wen2022fishing}, making the aggregate update the same as an individual update of a target user~\cite{pasquini2021eluding},
or looking to solve a blind source separation problem~\cite{kariyappa2022cocktail}. However, these attacks still have limited power in the FL setting. 
The first approach can only steal a single user image for each training round while additionally requiring multiple iterations prior to setup the attack. The second approach can use either analytic attacks or optimization attacks as a backbone but suffers from a similar scale limitation, only reaching a single client each round. The third approach can work with up to $1024$ images in aggregation for FedSGD (\# images/client $\times$ \# clients), but fails for more images.
{\em {\bf Thus, no prior method is able to scale to attacking multiple clients in FedAVG with secure aggregation.}}

\noindent \textbf{Our work.} We introduce \name\footnote{\namemeaning}, an attack whereby a malicious server can directly reconstruct the training data of multiple users in a single iteration using only the aggregate update. 
The attack works for both FedAVG and FedSGD, and even when secure aggregation is applied with hundreds of clients participating in the training round. To achieve this, the malicious server modifies the model that it sends to each target client. {\em The key insight behind our attack is that the server sends customized convolutional kernels to each client, so that the gradients of the input data of each client is separable in the aggregated update and is thus recoverable at the server.} With this separated weight gradient, an FC layer can then be used to leak the data of each individual client. Furthermore, the attack can increase the leakage rate by observing and adjusting the neurons in the FC layer that client images activate in each training round.  
This ability to leak client data regardless of aggregation breaks the previous scaling limitations of reconstruction attacks. While previous linear layer leakage methods~\cite{fowl2022robbing,boenisch2021curious} 
must scale the FC layer to address an increase in the the number of samples coming from an increasing batch size or increasing number of clients, we instead introduce a {\em split scaling} for this through our design. \name can scale to larger client dataset sizes by increasing the FC layer size, but it can also scale to a larger number of clients by increasing the number of convolutional kernels. 
This prevents a diminishing return in the leakage rate when higher numbers of clients are aggregated~\cite{boenisch2021curious}. 
With this property, we work especially well in large-scale aggregation such as cross-device FL, being able to leak $76\%$-$86\%$ of all images through only a single training round of FedAVG. 
On top of being able to break aggregation, gradients used for reconstruction also directly trace images back to the client that owns them. As a result, information on data ownership is also obtained and allows the attacking server another degree of freedom: the ability to target only high-value clients and identify their reconstructed data afterwards. 

Our main contributions are:

    \noindent (1) We introduce \name, an attack that allows data leakage even with secure aggregation. Using a single FedAVG training round with $100$ clients and a local dataset size of $64$, we leak $82.66\%$ ($5290$ of $6400$) training images from the aggregated update on CIFAR-100. 
    Further, \name can pinpoint which client each training sample comes from. \\
    (2) The attack works regardless of the number of clients in FedAVG and FedSGD. By increasing the size of the network and our split scaling technique, we can continue to scale our attack to increasing batch sizes {\em and} number of clients. \\
    (3) Using the convolutional scaling factor, \name is able to prevent images between separate local iterations from activating the same neuron, a fundamental problem in linear layer leakage for FedAVG. 
     This allows the attack to achieve a higher leakage rate for FedAVG than any prior linear layer attacks. \\
     (4) \name is able to handle non-IID clients by learning the distribution of the dataset and individual clients over multiple rounds. Through learning over just a single training round, we achieve up to a $62.3\%$ increase in leakage rate from the no-learning case on OrganAMNIST. 

\label{sec:introduction}

\vspace{-1mm}
\section{Background and related work}
\vspace{-2mm}

\begin{table*}
\begin{center}
\includegraphics[width=1.0\textwidth]{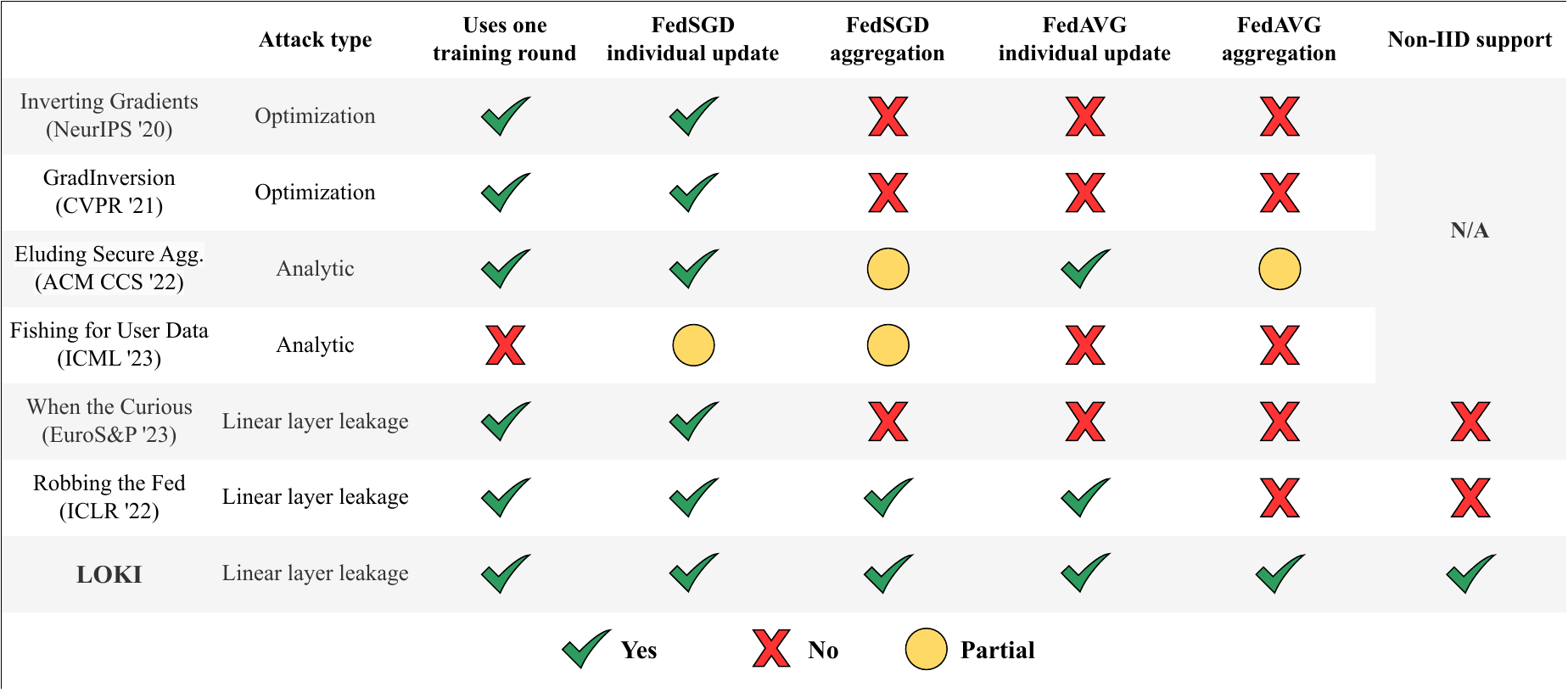}
\end{center}
\vspace*{-2mm}
\caption{\label{tab:comparison_table} Comparison of data reconstruction attack features. Partial support indicates the attack has limited reach/scale.}
\vspace*{-6mm} 
\end{table*}


While many different attacks on FL have been proposed, we focus on data reconstruction attacks, the strongest attacks on privacy in FL. 
Reconstruction attacks aim to break the fundamental notion of privacy of FL by obtaining client data directly through their updates. 
Prior work has done this through optimization~\cite{zhu2019deep,geiping2020inverting,zhao2020idlg,yin2021see,dimitrov2022data}, analytic attacks~\cite{pasquini2021eluding,wen2022fishing}, linear layer leakage~\cite{boenisch2021curious,fowl2022robbing}, or other approaches including GANs~\cite{hitaj2017deep,wang2019beyond}. These attacks typically aim to attack either individual client gradients or aggregated gradients. 
The following subsections will discuss the details and limitations of prior attacks.

\vspace{-1mm}
\subsection{Optimization-based attacks}
\vspace{-2mm}
Optimization approaches have shown great success in leaking data from individual updates, especially with smaller batch sizes. These attacks typically operate under the threat model of an honest-but-curious server or an external attacker that has access to the model and individual gradients from each client. With only this information, the attacker initializes some dummy data and computes the gradient of that data on the model.
\begin{equation}\label{eq:optim}
    x^* = \arg \min_{x}||\nabla L(x, y, \theta)-\nabla W||_2
\end{equation}
\noindent
An optimizer minimizes the difference between the generated gradient $\nabla L(x, y, \theta)$ and the ground truth gradient $\nabla W$ (benign client update). Here $x$ is the dummy data, $x^*$ is the reconstructed data, $L$ is the loss function, $y$ is the label, and $\theta$ is the model parameters.

More recent optimization approaches~\cite{geiping2020inverting,yin2021see} work under the assumption that user labels are known prior to optimization. Typically, these labels are directly retrieved through a zero-shot solution without using optimization approaches~\cite{zhao2020idlg,yin2021see}. Furthermore, regularizers and strong image priors specific to image data are often used to guide optimization results~\cite{geiping2020inverting,yin2021see}. These can also result in image artifacts typical of an image class, but not in the actual training image. These approaches have shown surprising success with image data on smaller batch sizes. However, as batch sizes increase, the fraction of images recovered decreases along with the reconstruction quality and the number of iterations required for the optimization also increases. One reason stated by~\cite{zhu2019deep} was that regardless of the order of images in the batch, the gradient will remain the same. Having multiple possible permutations then makes the optimization more difficult.
Another fundamental reason is that a larger batch size means more images and more variables for optimization.

\vspace{-1mm}
\subsection{Aggregate attack methods}
\vspace{-2mm}
There have been several attacks aimed specifically for aggregate gradients, however, they are currently limited in the attack scalability. 

In~\cite{pasquini2021eluding}, attackers send different models to clients such that the resulting aggregated gradient is only a targeted client's individual gradient. This is done by sending model parameters such that ReLU activated layers would have fully dead units (and an update with zero gradients) for any non-targeted client. The targeted client would get an attacked model, which would then be the only one to return non-zero gradients. On the other hand,~\cite{wen2022fishing} focuses on attacking a single data point through gradient magnification. The server sends weights to magnify the gradients of a targeted class by decreasing the model confidence on that class' prediction. Within the targeted class, the server will also focus on a specific feature in order to target a single sample. The resulting gradient will be similar to the gradient for a single image, allowing optimization-based approaches to retrieve the input. However, this process requires multiple training iterations. The attack can also only target a single image each time, and even in this case it does not succeed every time. Another method treats the inputs to a fully-connected layer as a blind source separation problem~\cite{kariyappa2022cocktail} where the weight gradients for the neuron make up a set of weighted combinations of the inputs. While the approach is able to attack aggregated gradients in FedSGD, the number of inputs is limited to be $1024$ or fewer. 

Another work~\cite{lam2021gradient} has looked to enable prior individual gradient methods through gradient disaggregation, separating out individual updates over time. However, along with requiring additional side-channel information about client participation, the method also requires a large number of training iterations to accomplish the goal. Similarly, due to partial user selection~\cite{cho2020client,chen2020optimal}, the server can reconstruct the individual models of some users using the aggregated models from previous rounds~\cite{pejo2020quality,secagg_so2021securing}. These approaches also require multiple training rounds and can be prevented by proper client selection so that no individual updates can be singled out.

\vspace{-1mm}
\subsection{Linear layer leakage attacks}
\begin{figure}[!t]
\begin{center}
\includegraphics[width=1.0\columnwidth]{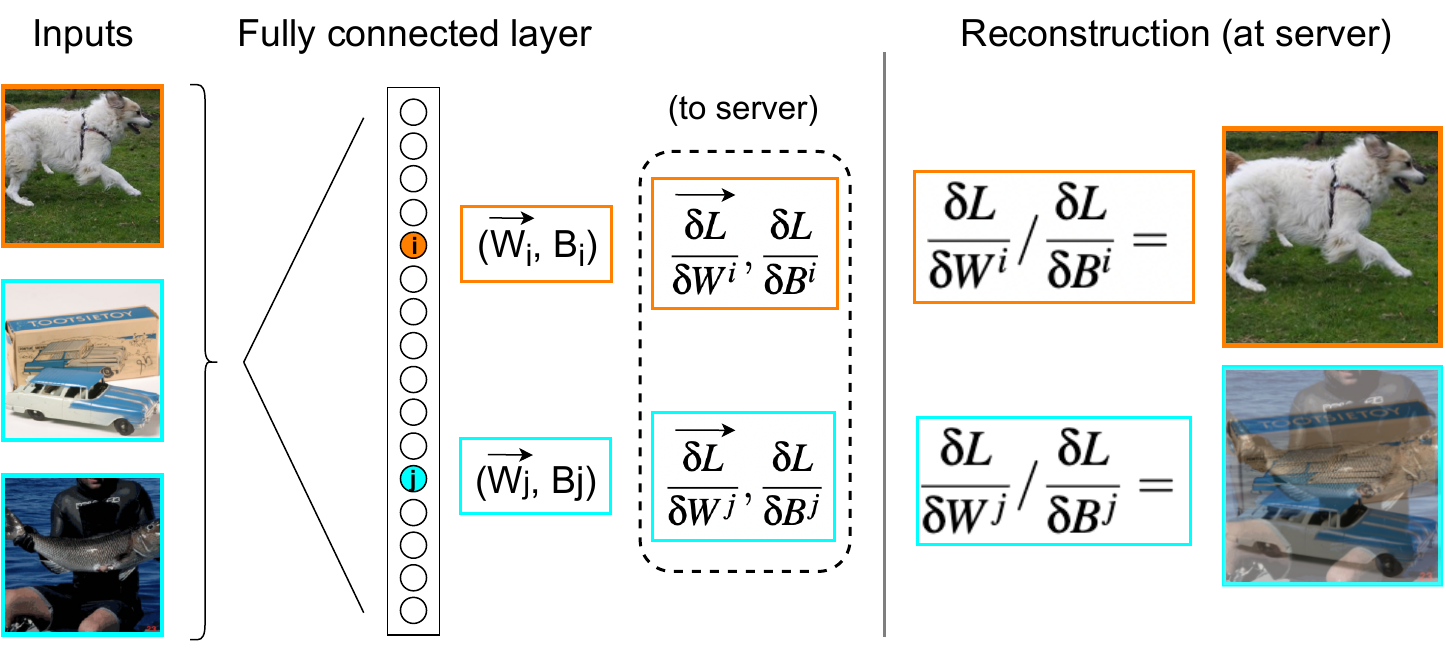}
\end{center}
\vspace*{-4mm}
\caption{\label{fig:linear-leak-method} Using the weight gradient $\frac{\delta L}{\delta W}$ and bias gradient $\frac{\delta L}{\delta B}$ of a fully connected layer to reconstruct the inputs. Neuron $i$ is only activated by a single image, while $j$ is activated by two. As a result, the reconstruction of neuron $i$ is correct while $j$ is a combination of images.}
\vspace*{-4mm}
\end{figure}

\vspace{-2mm}
Linear layer leakage attacks are a sub-class of analytic attacks that modify FC layers to leak inputs. Using the weight and bias gradients of an FC layer to leak inputs was discussed in~\cite{phong2017privacy,fan2020rethinking}. 
When only a single image activates a neuron in a fully connected layer, the input to that layer can be directly computed using the resulting gradients as
\begin{equation}\label{eq:1}
    x^i = \frac{\delta L}{\delta W^i} / \frac{\delta L}{\delta B^i}
\end{equation}
\noindent
where $i$ is the activated neuron, $x^i$ is the input that activates neuron $i$, and $\frac{\delta L}{\delta W^i}$, $\frac{\delta L}{\delta B^i}$ are the weight gradient and bias gradient of the neuron respectively. This idea forms the basis for several reconstruction attacks~\cite{boenisch2021curious,fowl2022robbing}. Figure~\ref{fig:linear-leak-method} shows the basic process of leaking images through an FC layer. 

When the fully-connected layer is placed at the start of a network, the data reconstructed from the the layer would be the input data. This reconstruction is exact, as opposed to the optimization approaches which function as estimations. However, inputs are only reconstructed exactly when a single data sample activates that neuron. If more than one input activates the neuron, the weight and bias gradients of these inputs will contribute to the batch gradient. 
When the gradient division of Equation~\ref{eq:1} is done to retrieve the input, the resulting reconstruction would be a combination of all contributing images, a case of failed attack. 

To alleviate this problem,~\cite{fowl2022robbing,boenisch2021curious} use malicious modification of the parameters in the FC layer. For~\cite{boenisch2021curious}, trap weights were introduced, initializing the weights randomly to be half positive, half negative. In order to ensure that neuron activation is less common, the negative weights come from a larger negative magnitude range than the positive weights. They also discuss the use of convolutional layers to push the input image forward, allowing the attack to function on models starting with convolutional layers followed by fully-connected layers. However, one of the main problems of the method lies with scalability. Even if the size of the FC layer increases proportionately with an increasing total number of images, the leakage rate decreases. On the other hand, Robbing the Fed (RtF)~\cite{fowl2022robbing} introduced another approach with higher leakage rate called ``binning", where the weights of the FC layer would measure some known continuous CDF of the input data such as image brightness. The bias for each neuron then serves as a different cutoff, allowing only inputs with a high enough value to activate it. The goal of this method would be that only one input activates each ``bin", where the bin is defined as the activated neuron with the largest cutoff (for ReLU, the largest negative bias)\footnote{The bin biases are set as negative. The weights are positive and so the negative bins are used to prevent ReLU activation.}. For any case where only one input activates a bin, it can then be reconstructed as
\begin{equation}\label{eq:2}
    x^i = (\frac{\delta L}{\delta W^i} - \frac{\delta L}{\delta W^{i+1}}) / (\frac{\delta L}{\delta B^i} - \frac{\delta L}{\delta B^{i+1}})
\end{equation}
\noindent
where $i$ is the activated bin and $i+1$ is the bin with the next higher cutoff bias. 

For Equation~\ref{eq:2} to hold true, the attack requires the use of two consecutive FC layers. 
The first layer is used to leak the inputs using Equation~\ref{eq:2} and the second FC layer maintains the requirement that $\frac{\delta L}{\delta B^i}$ and $\frac{\delta L}{\delta W^i}$ are the same for any neuron that the same input activates. This is achieved by having the same weight parameters connecting each neuron of the first FC layer to the second FC layer. For example, if the first FC layer has $1024$ units and the second has $256$, the weights connecting them would have 
a dimension of $1024\times256$. The above property indicates that every row of the weight matrix is equivalent, e.g. $[0, :]=[1, :] = \dots = [1023, :]$. 

\noindent \textbf{FedAVG}. While the previous method works for FedSGD, for FedAVG, the model changes during local iterations and this prevents the reconstruction attack. As a result,~\cite{fowl2022robbing} proposed the sparse variant of the attack which uses an activation function with a double-sided threshold (e.g., Hardtanh) such as:
\begin{equation}\label{eq:act_func}
    f(x) = \protect\begin{cases} 0 & x \leq 0 \\ 
    x & 0 \leq x \leq 1 \\ 
    1 & 1 \leq x  \protect\end{cases}
\end{equation}
\noindent With this activation function, only when the input is between $0$ and $1$ will there be a non-zero gradient. 

Using this activation function, neuron activation will be sparse (i.e., images will only activate a single neuron). However, this range between 0 and 1 for the non-zero gradient is fixed for all neurons. Since RtF's approach sets up neuron biases following a distribution of the images, the weights and biases of the FC layer will need to be adjusted to follow the new non-zero gradient range. This requires scaling the magnitude of these parameters based on the distance between the subsequent neuron biases. Consider that the weights originally measure average pixel brightness. In this case, all the weights would originally be set to $\frac{1}{N}$, where $N$ is the total number of pixels. Then, the weights and biases are rescaled as:
\begin{equation}\label{eq:param_scale}
\begin{split}
W^*_i = \frac{W_i}{b_{i+1} - b_i} \text{  } , \text{  } b^*_i = \frac{b_i}{b_{i+1} - b_i}
\end{split}
\end{equation}
\noindent where $W^*_i$ and $b^*_i$ are the scaled weights and biases of neuron $i$ respectively and $b_{i+1} - b_i$ is the distance between adjacent biases in the original distribution. This process uses the same distribution as the FedSGD case to setup the initial biases, while incorporating the fixed range of the new activation function by scaling the parameters. In FedAVG, after the clients send the updated model parameters, the server computes a ``gradient" as:
\begin{equation}\label{eq:fedavg_grad}
    \nabla W_{FedAVG} = \Theta_{t+1} - \Theta_{t} 
\end{equation}
where $\Theta$ is the model parameters for the securely aggregated model and $\nabla W_{FedAVG}$ the computed gradient. 
\label{sec:background}

\vspace{-1mm}
\section{Methodology}
\vspace{-2mm}
\begin{figure*}[t!]
\begin{center}
\includegraphics[width=1.0\textwidth]{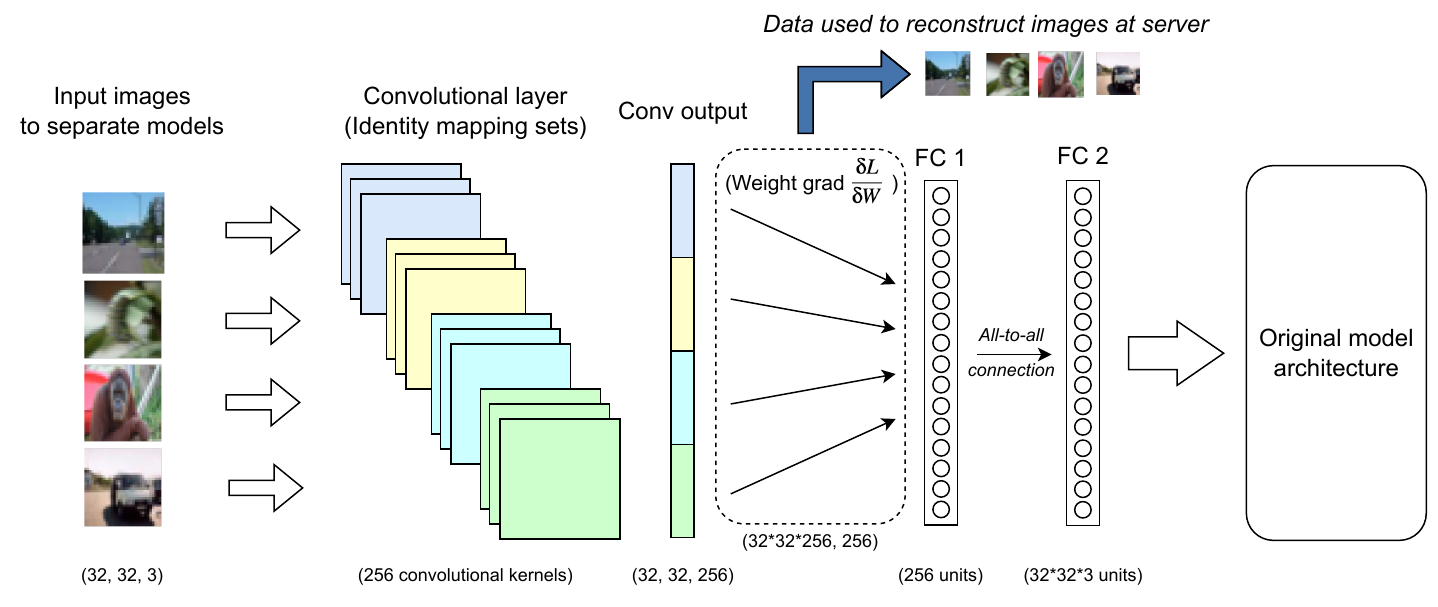}
\end{center}
\vspace*{-5mm}
\caption{\label{fig:identity_maps} The inserted attack module of a convolutional layer and two FC layers. Images are leaked using the gradients of the weight parameters connecting the convolutional output to the first FC layer. Weight gradient de-aggregation is done through separate identity mapping sets, indicated by the different color values in the convolutional kernels. In the aggregate update, the weight gradients of different clients are separated and used to leak the images. \iffalse \saurabh{Can we show within this figure, toward the bottom, the calculation for how many clients and batch size can be supported, i.e., calculate the sizes of our layers as a function of that.}\fi}
\vspace*{-3mm}
\end{figure*}
 
 As discussed previously, optimization becomes much more difficult with larger batch sizes and aggregation. The main problem is with more data to approximate, reconstructions end up having much lower quality and ultimately fails at larger batch sizes, even before aggregation is applied. On the other hand, the linear layer leakage methods~\cite{phong2017privacy,fowl2022robbing,boenisch2021curious} provide a powerful way of directly reconstructing training data without having reconstruction quality issues. Increasing the FC layer proportional to the number of total images in aggregation also allows~\cite{fowl2022robbing} to scale in FedSGD without losing effectiveness. However, the attack fails at scale for the much more difficult, and much more common, aggregation strategy of FedAVG. 


\subsection{Model assumptions and attack scope}
\vspace{-1mm}
\noindent \textbf{Threat model}. We operate under the same threat model as~\cite{fowl2022robbing,boenisch2021curious}, 
a malicious federated learning server with the ability to manipulate the model architecture and parameters before sending the model to clients. The server launches the attack by inserting a malicious module into the architecture, where the goal of the malicious server is to recover private training data. 

\noindent \textbf{System model}. We operate in cross-device FL, a setting where hundreds of clients participate in a single round of aggregation. 
The updates from the clients are aggregated through a secure aggregation mechanism before being made available to the server. 
Clients in this setting do not have the power to do a thorough verification of the models sent by the server. However, due to the application of secure aggregation, outside of the aggregate update, the server cannot receive any information about the clients or about individual updates. 

\noindent \textbf{Attack scope}. In this work, we target both FedAVG and FedSGD. Under the threat model, \name can leak the inputs to that model
regardless of the data type (e.g. image, video, audio, text), similar to~\cite{fowl2022robbing}. We experimentally demonstrate this attack on image data. 

\subsection{Scalability problem}
\noindent \textbf{FedAVG}. Since Robbing the Fed (RtF)~\cite{fowl2022robbing} can scale to aggregation in FedSGD by increasing the size of the FC layer, it is not immediately apparent why the same cannot be done in FedAVG with the sparse variant. In theory, the images will activate the same set of neurons between FedAVG and FedSGD (given a bit of difference coming from parameters changing slightly during local iterations), so reconstruction rate will be the same. However, there is a critical problem with scaling the size of the FC layer in FedAVG. The biases will follow the same distribution regardless of the size of the layer. If the distribution is the average pixel brightness, the range of values will always be between $0$ and $1$. Having a larger FC layer size results in more neurons in the same range, and the distance between adjacent biases decreases. This means the value of $b_{i+1} - b_i$ gets smaller and as a result, the scaled weights $W^*_i$ and biases $b^*_i$ become larger (Equation~\ref{eq:param_scale}).
\begin{equation}\label{eq:fedavg_predicion_update}
\begin{gathered}
||W_t, b_t|| \gg ||\nabla_{W_t}, \nabla_{b_t}||\\
W_{t+1} = W_t + \alpha\cdot\nabla_{W_t} \approx W_t\\
b_{t+1} = b_t + \alpha\cdot \nabla_{b_t} \approx b_t
\end{gathered}
\end{equation}
The increasing size of the FC layer results in increasing $W^*_i$ and $b^*_i$ parameter magnitude, 
but the magnitude of each client's gradient update $(\nabla_{W_t}, \nabla_{b_t})$ will remain relatively unchanged. 
As a result, the relative parameter shift in the model between different local iterations becomes much smaller. 
This same property is the Achilles heel, as once the magnitude of the parameters increase beyond a certain point, the update computed for the FC layer essentially becomes zero, i.e., $W_{t+1}-W_{t}=0$ for the FC layer (Equation~\ref{eq:fedavg_predicion_update}). 
Even before the updates become zero, the values lose precision which reflects in poor reconstruction quality. This becomes a large problem, resulting in a heavily decreasing leakage rate and reconstruction quality as the number of clients in aggregation increases.

The default precision used for deep learning models in most packages is floating-point 32 (FP32). If the original value of the parameters is much larger than the update, adding the gradient to the parameters results in a loss of precision or in the extreme case, no update at all. This is typically caused by increasing the FC layer size, but also can be affected by other factors such as the learning rate or local mini-batch size. RtF~\cite{fowl2022robbing} mentions a way to increase the effectiveness of the method by scaling the linear distribution $h_{new}(x) = c\cdot h_{original}(x), c > 1$
to minimize the change in the parameters over the local iterations. However, what we find is that this does not address the problem in FedAVG, as scaling the distribution not only scales the weights of the FC layer, but also the biases. Consider a case where the image brightness distribution is scaled by a factor of 10. Since the weights used to measure it increase by $10\times$ to achieve this, the biases used as cutoffs also have to be scaled by the same factor. These layer parameters are then scaled by the distance between the biases $b_{i+1} - b_i$, which in turn is also $10\times$ larger. This ultimately results in no change in the values of $W^*_i$ and $b^*_i$ of Equation~\ref{eq:param_scale}. 
Thus, scaling the linear distribution does {\em not} fix the precision problem.

\noindent \textbf{FedSGD}. For FedSGD on a single client, if multiple images from that client's batch activate the same neuron, reconstruction fails. This problem is directly exacerbated with aggregation. If multiple images \textit{across any of the clients} activate the same neuron, 
reconstruction at the server also fails. For a successful reconstruction, only one image across all client batches can activate the same neuron. Consider a case where we have $100$ clients each with a batch size of $64$. If we have an FC layer of $256$ units, this means $64\cdot100=6400$ images are shared, resulting in an average of $25$ images per neuron. For linear layer leakage, the generalization to these larger-scale attacks is done by simply increasing the size of the FC layer. In the case of RtF~\cite{fowl2022robbing}, a proportional increase in FC layer size with the number of images will maintain the same leakage rate. However, for trap weights~\cite{boenisch2021curious} the leakage rate will still decrease.

Therefore, while the underlying reasoning is different for the sparse variant of RtF~\cite{fowl2022robbing} in FedAVG and for trap weights~\cite{boenisch2021curious} in FedSGD, the fundamental scalability problem is the same. \textit{Current linear layer leakage methods only scale the size of the FC layer for an increasing batch size or number of clients, which results in scalability problems.} \name breaks this scaling problem for both settings and separates the scaling of the batch size and the number of clients. We increase the size of the FC layer for larger dataset sizes (batch sizes) and the number of convolutional kernels for more clients. Furthermore, we introduce a convolutional scaling factor that prevents multiple activations of a neuron across epochs and helps mitigate precision problems coming from factors such as local mini-batch size or learning rate.

\subsection{Attack architecture}
\vspace{-1mm}
We insert an attack module at the start of a model that consists of a convolutional layer followed by two FC layers. This module is shown in Figure~\ref{fig:identity_maps}. 
We leak images using the computed gradients of the weight parameters connecting the output of the convolutional layer to the first FC layer. The dimension of the output of the convolutional layer depends on the image size and number of kernels. For a $32\times32$ image and $100$ kernels, the output would have dimension $32\times32\times100$.

The size of the first FC layer depends on the local dataset size for FedAVG or batch size for FedSGD. Generally, we add $4$ units in the layer for each image. With a batch size of $64$, this would be $256$ units. Every unit in the first FC layer is connected to the second FC layer, which is the input to the rest of the model architecture and will have the same dimensions as the input image. As an inserted module, it is important that the input and output dimensions are the same, so that the dimensionality expected by the benign model is not altered.

\vspace{-1mm}
\subsection{Convolutional parameters}
\vspace{-2mm}
We start our discussion by describing the attack in FedSGD before discussing the (significant) sophistication for FedAVG. Previous works~\cite{fowl2022robbing, boenisch2021curious} have discussed the idea of leaking with convolutional layers followed by FC layers. The standard way to achieve this would be the use of convolutional kernels to push the image forward and have FC layers further in the model leak the inputs. We will further explore the use of convolutional parameters.

\begin{figure}[!t]
\begin{center}
\includegraphics[width=1.0\columnwidth]{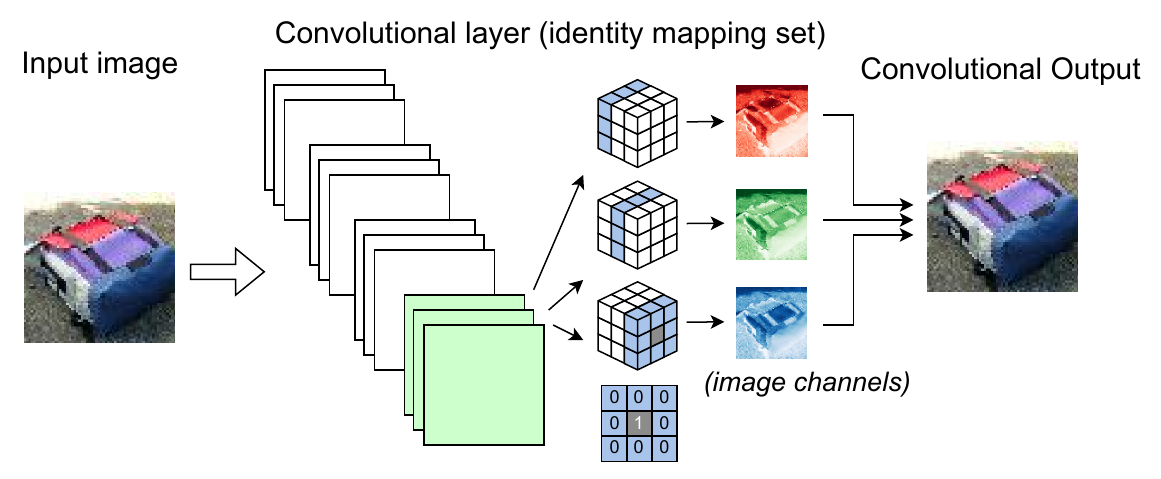}
\end{center}
\vspace*{-5mm}
\caption{\label{fig:conv-sets} Identity mapping set for a 3-channel input. The first three kernels ($3\times3\times3$ cubes) of the convolutional layer push a different input channel forward. All parameters are zero except for a single element. The 2D slice with a single non-zero value is shown in the figure, and the locations of the slice for each kernel allows them to push different input channels forward.}
\vspace*{-5mm}
\end{figure}

For a 3-channel input image (RGB), pushing the image forward can be done with three kernels in a convolutional layer. 
If we have $3\times3$ kernels, the dimension of each kernel would be $3\times3\times3$ with the final dimension corresponding to the input channels. 
For any given client, we will need a minimum of three kernels for a 3-channel image. Within each kernel, there is only one key channel, which will have a  value of 1 in the center and all other elements will be zero. The three kernels will have this in a different channel. We call the single non-zero value the key value $kv$. When the convolutional kernels are applied to the RGB input image, each kernel will push a separate channel forward, resulting in the image in its entirety being pushed through. 
This setup is shown in Figure~\ref{fig:conv-sets}. We will call these three convolutional kernels an \textit{identity mapping set}. This operation only requires the use of three convolutional kernels, a small number in the context of typical models. The other kernels are not needed to propagate any information, so the outputs are set to zero. There are multiple ways to set the kernel parameters such that the output is zero, but generally a large negative bias or zero/negative weights can cause the output to be zero. 

This simple approach still severely under-utilizes the number of convolutional kernels. Consider a situation where we have 256 convolutional kernels. To push one 3-channel image forward, we only use three kernels. As a result, the other 253 kernels do not contribute. With this in mind, we propose the use of multiple, separate identity mapping sets. Each set requires three kernels (number of input channels), and following this, we have $\lfloor \frac{N}{3} \rfloor$ separate identity mapping sets, where $N$ is the total number of convolutional kernels. These separate identity mapping sets, each corresponding to a different set of three convolutional kernels, are then used in different models and are sent to different clients.

\vspace{-1mm}
\subsection{De-aggregated update}
\label{sec:reconstruction_equation}
\vspace{-2mm}
By sending the carefully crafted separate convolutional kernels to each client, the weight gradient for each set of identity mapping sets is non-zero only for a different set of convolutional output channels. Therefore, when updates are aggregated together, the weight gradients remain separate. During the reconstruction phase, if inputs from different clients activate the same bin, 
the computed weight gradient of that bin will not be shared between clients. The only inputs that can share the same set of weight gradients would be images within that single client's batch. This essentially allows the size of the FC layer to scale only based on the client batch size, and {\em not with the number of clients}. 

When reconstructing images, this allows us to work with different sets of weight gradients, each corresponding to separate identity mapping sets (and client model). Instead of Equation~\ref{eq:2}, we would then reconstruct images as:
\begin{equation}\label{eq:3}
    x^i_{k} = (\frac{\delta L}{{\delta W^i}_k} - \frac{\delta L}{{\delta W^{i+1}_k}}) / (\frac{\delta L}{{\delta B^i}_k} - \frac{\delta L}{{\delta B^{i+1}}_k})
\end{equation}
\noindent
where $k$ indicates the client and the corresponding weight and bias gradients respective to their identity mapping set. $x^i_k$ is the input from client $k$ that activates bin $i$. Figure~\ref{fig:identity_maps} shows the process of using separate identity mapping sets to split the aggregate weight gradient. This allows the attacker to leak images from each client separately after aggregation.

This decoupled weight gradient partially solves the scaling problem of reconstructing images as the number of clients participating in aggregation increases. However,
Equation~\ref{eq:3} brings up another problem for reconstruction, as the bias gradient is needed for the computation. While the weight gradients are separated through the identity mapping sets, the bias gradients are not. 
Secure aggregation aggregates the bias gradients of each neuron (of the first FC layer). Thus the neuron $i$'s bias values from the FC layers of all clients are aggregated. Consider that an image $j$ activates neuron $i$ in client $k$ and another image $j'$ activates the neuron $i$ in client $k'$ (note that the neurons are physically separate before aggregation as they are in the local FCs of clients $k$ and $k'$). After secure aggregation, the bias update of neuron $i$ from clients $k$ and $k'$ are coupled and therefore the server is not able to use Equation~\ref{eq:3} to decouple images $j$ and $j'$. To solve this problem, we look at the purpose of the bias gradient in reconstruction.

\begin{figure}[t!]
\begin{center}     
\subfigure[Ground truth]{\label{fig:bright_shift_gt}\includegraphics[width=0.47\columnwidth]{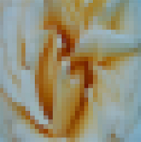}}\hspace{5mm}%
\subfigure[Reconstruction]{\label{fig:bright_shift_rec}\includegraphics[width=0.47\columnwidth]{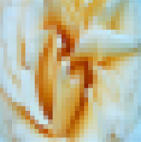}}\hspace{5mm}%
\end{center}
\vspace*{-3mm}
\caption{\label{fig:bright_shift} Image reconstructed using Equation~\ref{eq:fedsgd_recon} to scale the maximum to 1. Ground truth image (a) has a maximum intensity of 0.7804, resulting in a brightness shift in the reconstructed image (b).}
\vspace*{-5mm}
\end{figure}

Observing Equation~\ref{eq:3}, we note that for reconstruction of each neuron we subtract the weight gradients, and the resulting value is divided by the subtraction of bias gradients. Previously we mentioned how each neuron has a single bias, so after subtracting bias gradients, the resulting value remains a scalar. The purpose of the bias gradient, then, \emph{is to scale the value produced by subtracting the weight gradients such that it becomes the same as the input}. 
Therefore, as long as we know what the weight gradient needs to be scaled to, we will not need to know the exact bias gradient. Rather, knowledge of the input range is all that is required to reconstruct the input. For image datasets such as MNIST, CIFAR-10, or Imagenet, the training data will be between 0 and 1. 
Using this, the images can be reconstructed by scaling the gradient such that the maximum value is 1, without requiring knowledge of the bias gradient.
\begin{equation}\label{eq:fedsgd_recon}
    x^i_{k} = \frac{abs({\frac{\delta L}{\delta W^i}_k} - \frac{\delta L}{\delta W^{i+1}}_k)}{max(abs({\frac{\delta L}{\delta W^i}_k} - \frac{\delta L}{\delta W^{i+1}}_k))}
\end{equation}
\noindent
The numerator is the absolute value of the subtracted weight gradient for client $k$ between neuron $i$ and $i+1$, while the denominator is the maximum of that value across the set of three slices in the identity mapping set corresponding to client $k$. 
The technical question now is much simpler --- estimating the maximum value of the denominator. The input range can be estimated by the server, or may be known through other public sources, such as through the standardized normalization prior to training. This estimation could be imprecise, but we see empirically that inaccuracies in this estimation do not hurt the reconstruction performance much, as the image remains the same structurally. Figure~\ref{fig:bright_shift} shows a reconstructed image with a shifted brightness. 
The maximum pixel intensity of the ground truth was 0.7804 but was scaled to be 1.0 during reconstruction. Importantly, the error in reconstruction is contained to just that image and does not affect the reconstruction of successive images in that batch. 
Thus, our key result is that in linear layer leakage methods, the bias gradient is \textit{not} required for reconstruction of data.

\vspace{-1mm}
\subsection{FedAVG and convolutional scaling factor}
\label{sec:FedAVG_CSF}
\vspace{-2mm}
By itself, the trap weights attack~\cite{boenisch2021curious} suffers from scalability problems in FedSGD. However, utilizing the convolutional attack structure of \name, the scalability problems are fixed through the split scaling between the number of clients and batch size (Figure~\ref{fig:wtc_leakage}).
The methodology is generalizable to FedAVG by using a double-bounded activation function as in Equation~\ref{eq:act_func} for the FC layer and scaling the parameters using the distance between adjacent biases using Equation~\ref{eq:param_scale}. 
The split scaling design mitigates the scalability problems in FedAVG coming from precision when having large parameters and small gradients. Since the FC layer size does not scale based on the number of clients, the distance between the biases in the layer does not change. As a result, the magnitude of the parameters $W^*_i$ and $b^*_i$ will not increase, helping alleviate the problem.

However, factors such as the learning rate or local mini-batch size still affect the update (gradient) size. The learning rate directly affects the size of the gradient update and the local mini-batch size also affects the magnitude of the individual gradient contributions from each image. For example, for a mini-batch of 10 images the gradients are averaged over the 10 images. For 20 images the gradients are averaged over 20 images, so the individual contributions are smaller. Furthermore, the FC layer size still needs to increase linearly with the local dataset size (the total number of images used in training) in order to maintain the same leakage rate. 
These factors still impact the precision problem.

To address this, we introduce a convolutional scaling factor (\textit{CSF}) for the FedAVG attack. We can scale the image coming out of the convolutional layer by using a different key value ($kv$) for the identity mapping sets in the convolutional kernels. This can be used to offset the increase in the magnitude of the weight parameters $W^*_i$ coming from the increasing FC layer size or the small gradients coming from the learning rate or local mini-batch size. (Note that since we do not use the bias parameters $b^*_i$ to reconstruct images, we do not need to worry about update precision for them.) Using the \textit{CSF}, we can maintain the distribution and scale the parameters as:
\begin{equation}\label{eq:conv_scale_factor}
    kv_{new} = \textit{CSF}\cdot kv \text{  } , \text{  } W^*_{i,new} = \frac{1}{\textit{CSF}}\cdot W^*_i
\end{equation}
\noindent
where $kv$ is the non-zero value (typically 1) in the identity mapping set. The \textit{CSF} allows us to prevent precision problems during reconstruction, but also helps minimize the changes in the convolutional kernel parameters to preserve the original purpose of pushing the inputs forward. The value produced
after the image passes through the convolutional layer and
the weights of the FC layer remains the same as in the baseline ($\textit{CSF}=1$) in order to fit the distribution for the biases correctly.

Crucially, the \textit{CSF} is also used to help address the fundamental problem in linear layer leakage of having multiple images activate the same neuron. After an image activates a neuron during a local iteration, a smaller value of $W^*_{i,new}$ results in a larger relative change in the parameters. 
However, after an image activates the neuron, we do not want any additional activations in subsequent local iterations. 
Therefore, when the \textit{CSF} is large, the changes in the weights after each local iteration become large enough so additional images do not activate the same neuron in subsequent local iterations (if images are orthogonal, e.g., their dot product is zero, there can still be an activation of the same neuron. However this is a very rare case in practice). When multiple images in separate local iterations would have activated the same neuron, only the first image ends up actually activating it due to parameter shift. This method allows for a correct reconstruction of the first image, which was previously impossible. As a result, \name achieves higher leakage rate in the FedAVG attack compared to FedSGD. This same property of the \textit{CSF} also prevents images from different local epochs from activating the same neurons.
\label{sec:methodology}

\vspace{-1mm}
\section{Additional attack details}
\vspace{-2mm}

\subsection{Setting up FC layer biases}
\label{sec:setting_biases}
\vspace{-1mm}
While leaking images from a linear layer is superior in reconstruction speed and quality compared to optimization, the initialization of neuron biases can pose a challenging problem in practice. The weights of the FC layer measure some aspect of an image such as the average pixel intensity and the biases must be used as cut-offs for image activations. If not set properly, the number of recovered inputs is lower. For example, if we initialize completely randomly on CIFAR-100, the leakage rate can drop by over 20$\%$ as shown in Appendix Table~\ref{tab:bias_initialization}.
If the server knows the dataset distribution for average pixel intensity, as assumed in~\cite{fowl2022robbing}, setting the biases of the FC layer is simple. However, in practice client datasets are private, so the server is unlikely to have such knowledge.

We therefore incrementally learn the distributions of the dataset.
If the FC layer weights were used to measure the average pixel intensity, this value must be between $0$ and $1$. An initialization for the biases following this could be so that the biases have a mean of $-0.5$ and standard deviation of $0.25$.
This setting is progressively improved through training iterations by observing the neuron activations. After receiving the aggregated gradient, the server observes which neuron activations result in a successful reconstruction and which do not (same neuron activated multiple times). For each observation, it also notes the bias values of the neuron and using these observed biases, it computes a new mean and variance over them to use for the initialization of the biases for the next iteration. Over several training rounds, the server arrives at a close estimate of the dataset distribution without any prior knowledge. 
This determination is done by the server separately for each client in the case of non-IID data where clients may have vastly different distributions. Additionally, the server can observe the type of distribution (i.e. normal, multi-Gaussian etc.) and set up biases to fit them. For this work we setup biases following a normal distribution. 

\vspace{-1mm}
\subsection{Identifying client data}
\vspace{-2mm}
Even if the aggregated gradient is able to leak training data, it may appear to be of some comfort that the attacker cannot identify which data belongs to which client. However, \name gives the attacker the ability to identify the owner of the reconstructed data even through aggregation.
The identity mapping sets keep the weight gradients for each client separate after aggregation. As a result, when reconstructing inputs, the set of weight gradients used allows the server to identify which model it originates from and hence, which client it originates from. 
With this information, the server may subsequently focus on specific high-value clients.

\vspace{-1mm}
\subsection{Parameter comparison}
\vspace{-2mm}
For linear layer leakage, attacks where the server modifies the model, a key question is how many parameters the attack adds. A small addition is desirable as we are operating in the cross-device setting where communication and storage are at a premium. The  fundamental premise behind linear layer leakage is using the gradients of a model update to store the information of the input data. For example, with a $32\times32\times3=3072$ image, the model must have at least $3072$ weight parameters to store this. To store $1000$ images, we would then need $3072\cdot1000=3,072,000$ parameters. However, exactly $1000$ units for $1000$ images assumes that every image activates a separate neuron, an extremely optimistic case (for the attack). The FC layer size will typically need to be larger than the number of images (4$\times$ is typical).

For comparison between \name and RtF for number of added parameters, consider we have $100$ clients with a local dataset size of $64$. Each image is given $4$ units, resulting in $256$ units for each client. For simplicity, we will ignore the number of bias parameters, as the amount is much smaller than the weight parameters (e.g. for \name, $0.005\%$ of the total parameters are biases). For RtF~\cite{fowl2022robbing}, it ends up needing 157M parameters while \name needs roughly half (50.7\%). The basic reason for this reduction is that RtF needs a large first FC layer (25,600 = batch size $\times$ multiplicative factor (4) $\times$ \# clients), while we only need 256. The number of connections from the first to the second FC layer are determined by dimensionality considerations and cannot be reduced. This improvement in \name is taking into account the additional parameters due to the convolutional kernels. 

The added parameters are also extremely sparse, as only the weight parameters connecting the identity mapping set kernel outputs to the first FC layer need to be non-zero ($3$ out of $300$). When attacking $100$ clients, 98\% of our total added parameters are zero. As the number of clients increases, the absolute number of non-zero parameters at each client stays constant. Sparsity can allow for additional compression of the model/update and more efficient optimization to reduce the computational and communication cost overhead significantly~\cite{duff1989sparse,paszke2019pytorch,abadi2016tensorflow}.

\subsection{Model inconsistency}
\label{sec:model_inconsistency}
\vspace{-2mm}
We have so far discussed the attack in the context of using separate identity mapping sets for each client. This method inherently creates model inconsistency among clients. However, this is not a requirement in \name. Some clients can be sent the same convolutional kernels. For any clients that are sent the same identity mapping set, if images between clients activate the same neuron, there will be failure in reconstruction, so the size of the FC layer will need to be increased accordingly. The two extremes of model inconsistency are: full model inconsistency where all clients get different models, and no model inconsistency where all clients get the same model. 

\begin{figure}[t!]
\vspace{-1mm}
\begin{center}
\includegraphics[width=0.9\columnwidth,trim={0mm 0mm 0mm 11mm},clip]{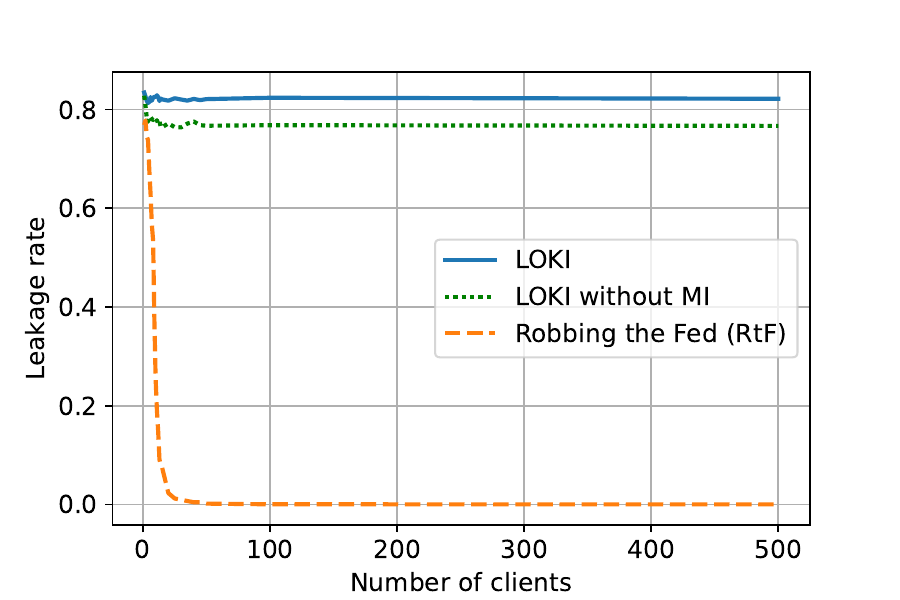}
\end{center}
\vspace*{-5mm}
\caption{\label{fig:scalability_rtf} Leakage rate of \name with and without model inconsistency (MI) and Robbing the Fed (RtF) for a number of clients between 1-500 on CIFAR-100. Clients train with 8 local iterations of mini-batch size 8.}
\vspace*{-2mm}
\end{figure}

In the latter case of no model inconsistency, the original scalability problem seems to exist. As the size of the FC layer increases with the total number of clients or local dataset size, the magnitude of the scaled parameters increases. However, the convolutional scaling factor of \name offsets this change and prevents reconstruction problems. The $\textit{CSF}>1$ setting allows us to avoid precision problems and thus keep leakage rate high. There are some additional downsides to not using model inconsistency for \name, such as a slightly lower leakage rate (Figure~\ref{fig:scalability_rtf}), larger model size overhead (Figure~\ref{fig:model-inconsistency-size}), or losing the ability to identify client data ownership. However, with no model inconsistency \name still achieves scalability to an arbitrary local dataset size or number of clients in FedAVG.

\label{sec:additional}

\vspace{-1mm}
\section{Experiments}
\vspace{-2mm}
\begin{figure*}[!ht]
\begin{center}     
\subfigure[Ground truth local dataset]{\label{fig:ground_truth_recon}\includegraphics[width=0.65\columnwidth]{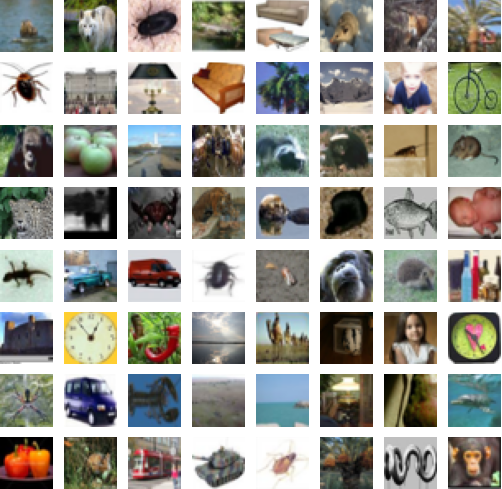}}\hspace{5mm}%
\subfigure[\name reconstruction]{\label{fig:mandrake_recon}\includegraphics[width=0.65\columnwidth]{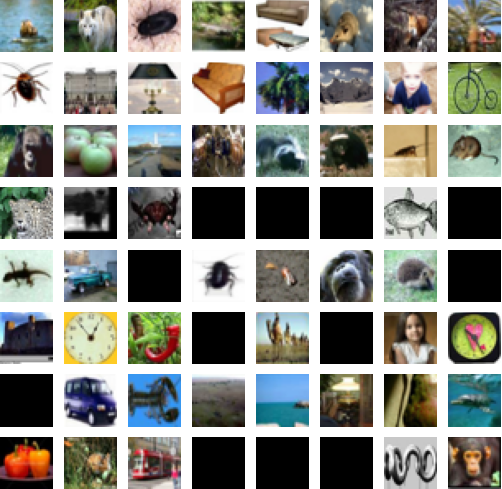}}\hspace{5mm}%
\subfigure[Robbing the Fed~\cite{fowl2022robbing} reconstruction]{\label{fig:rtf_recon}\includegraphics[width=0.65\columnwidth]{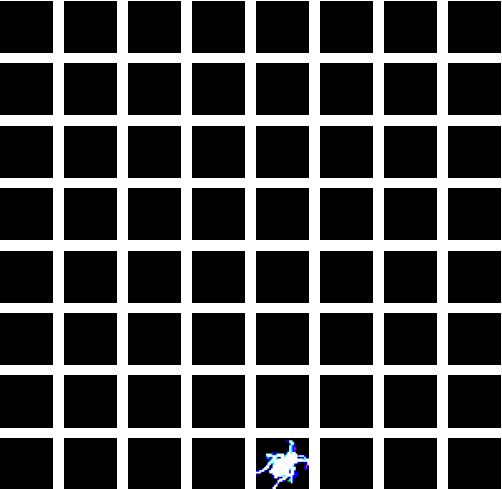}}
\end{center}
\vspace*{-3mm}
\caption{\label{fig:reconstruction_examples_cifar100} \name and Robbing the Fed (RtF) reconstructions on CIFAR-100 for 1 client out of 100 in FedAVG aggregation. Clients use 5 epochs with 8 local iterations and a local mini-batch size 8. Empty boxes indicate images were not reconstructed because multiple images activated a neuron. In the case of RtF, nearly all empty boxes occur due to precision problems causing the computed gradient to be zero rather than multiple activations.}
\vspace*{-3mm}
\end{figure*}

In this section, we provide experimental results for \name on aggregated updates in FedAVG. We assess our baseline attack on the CIFAR-100~\cite{krizhevsky2009learning}, Tiny ImageNet\cite{le2015tiny}, and MNIST~\cite{lecun1998mnist} datasets. The leakage module is used with a ResNet-50~\cite{he2016deep}, but the benign model itself does not affect reconstruction. The FC layer weights measure the average pixel intensity and the biases of the FC layer are set up according to the dataset distribution (known to the server or learned through the first few training iterations). Unless otherwise specified, we have 4 units in the FC layer per image for \name and Robbing the Fed (RtF)~\cite{fowl2022robbing}. 

We first show the leakage rate that \name achieves for each dataset in FedAVG before also showing the scalability in FedSGD. Further experiments show various FedAVG training settings, the effects of the CSF, a non-IID attack which we evaluate using the OrganAMNIST~\cite{medmnistv2} dataset (which has larger distribution differences when using a class-based non-IID skew compared to CIFAR-100), and differential privacy. Some additional experimental results are shown in Appendix~\ref{sec:appendix}.

\begin{table}[!t]
\small
\begin{center}
\begin{tabular}{|p{0.174\columnwidth}|c|c c|} 
\hline
\textbf{Dataset} & \textbf{Metrics} & \textbf{\name} & \multicolumn{1}{c|}{\begin{tabular}[c]{@{}c@{}}\textbf{RtF~\cite{fowl2022robbing}} \\ \textbf{+ MI~\cite{pasquini2021eluding}}\end{tabular}} \\
\hline
\multirow{3}{*}{\textbf{CIFAR-100}} & Leaked imgs & 5290 & 50 \\
& Total imgs & 6400 & 6400 \\
& \textbf{Leakage rate} & \textbf{82.66\%} & \textbf{0.78\%} \\
\hline
\multirow{3}{0.19\columnwidth}{\textbf{Tiny ImageNet}} & Leaked imgs & 5202 & 49 \\
& Total imgs & 6400 & 6400 \\
& \textbf{Leakage rate} & \textbf{81.28\%} & \textbf{0.77\%} \\
\hline
\multirow{3}{*}{\textbf{MNIST}} & Leaked imgs & 4907 & 49 \\
& Total imgs & 6400 & 6400 \\
& \textbf{Leakage rate} & \textbf{76.67\%} & \textbf{0.77\%} \\
\hline
\end{tabular}
\end{center}
\vspace*{-0mm}
\caption{\label{tab:leakage_table} Leakage rate for FedAVG aggregated update with 100 participating clients. $\alpha=1e-4$, $\textit{CSF}=100$ and 5 local epochs of 8 iterations of mini batch size 8 used. }
\vspace*{-8mm}
\end{table}

\vspace{-1mm}
\subsection{FedAVG aggregation attack}
\label{sec:fedavg_leakage_rate}
\vspace{-2mm}
Figure~\ref{fig:scalability_rtf} shows the leakage rate for RtF with a varying numbers of clients between 1-500 averaged over 10 runs. We define an image as leaked only if a single image activates the neuron and has a reasonable reconstruction quality with SSIM above 0.5. If two or more different images (not same image across multiple epochs) activate the neuron, even with a high SSIM, we do not count it as leaked. Clients are trained with with 8 local iterations with mini-batch size 8 on CIFAR-100, and learning rate $\alpha=1e-4$. We use \name with $\textit{CSF}=100$ ($\textit{CSF}=100\cdot\textit{num. of clients}$ when not using model inconsistency). While RtF is applicable to a smaller numbers of clients, the leakage rate very quickly decreases as the number of clients increases. With only 15 clients, the leakage rate of RtF drops to 7.57$\%$. With 30 clients, the leakage rate drops to 0.95\%. As the FC layer size increases with the number of clients, the leakage rate decreases due to precision problems.
We use a relatively small number of local iterations and mini-batch size, but if either increase, RtF will function on even fewer clients. \name is unaffected by training parameters (learning rate, local iterations, mini-batch size) and achieves a high leakage rate regardless of the number of clients. Without model inconsistency, \name has a slightly lower leakage rate but still achieves scalability with no impact on reconstruction quality.

\begin{figure*}[t!]
\begin{center}     
\subfigure[10 clients]{\label{fig:rtf_10cl}\includegraphics[width=0.65\columnwidth,trim={18mm 18mm 18mm 18mm},clip]{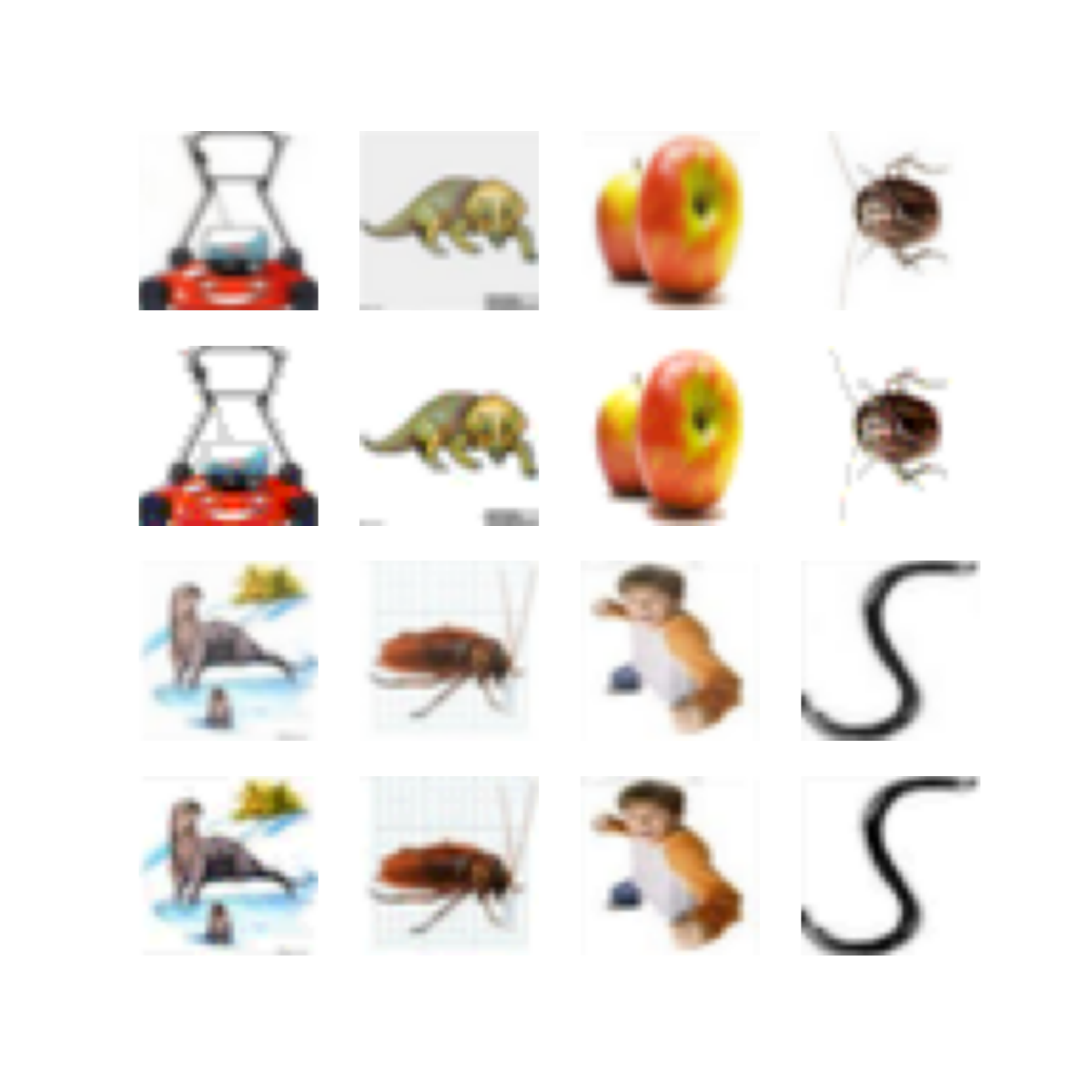}}\hspace{5mm}%
\subfigure[25 clients]{\label{fig:rtf_25cl}\includegraphics[width=0.65\columnwidth,trim={18mm 18mm 18mm 18mm},clip]{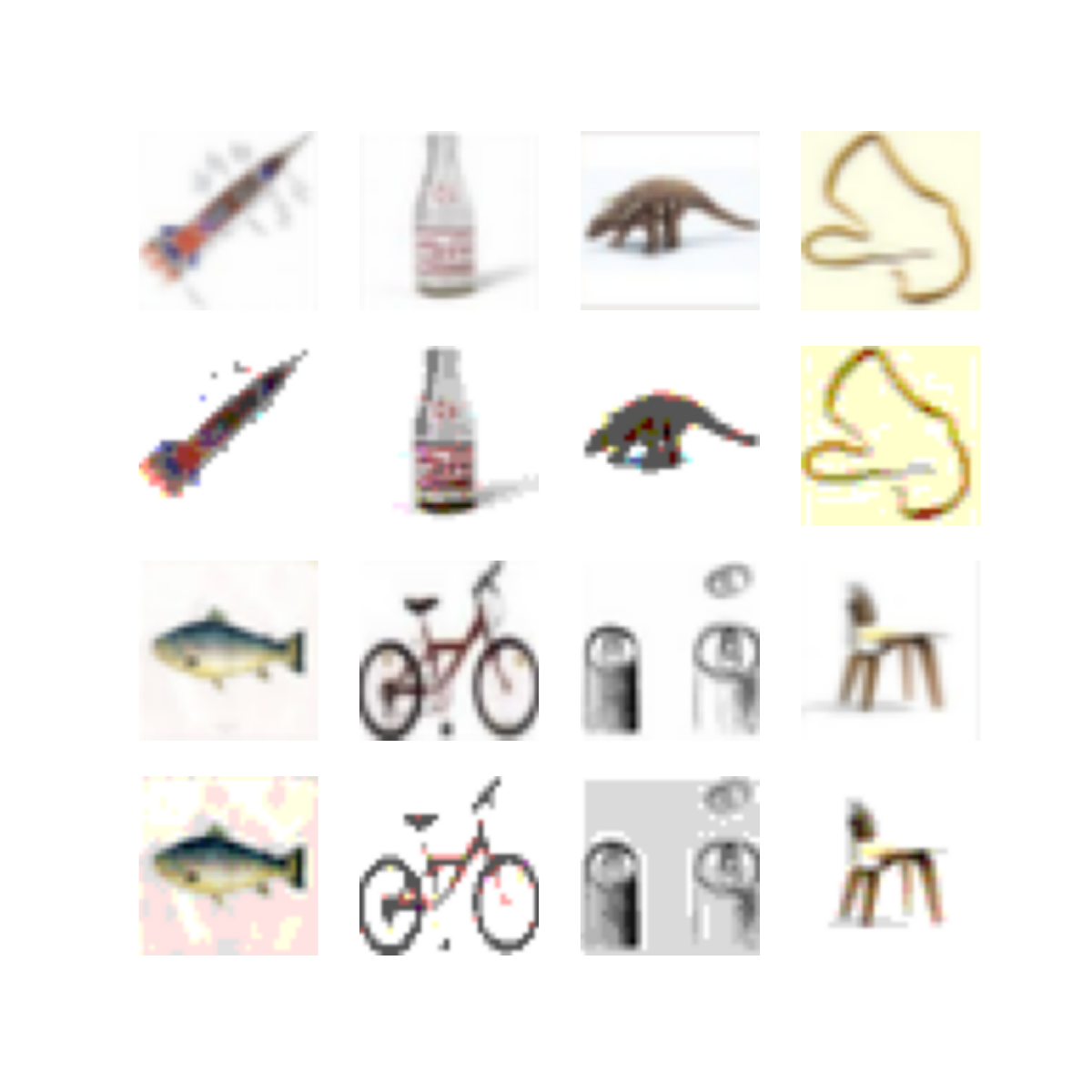}}\hspace{5mm}%
\subfigure[50 clients]{\label{fig:rtf_50cl}\includegraphics[width=0.65\columnwidth,trim={18mm 18mm 18mm 18mm},clip]{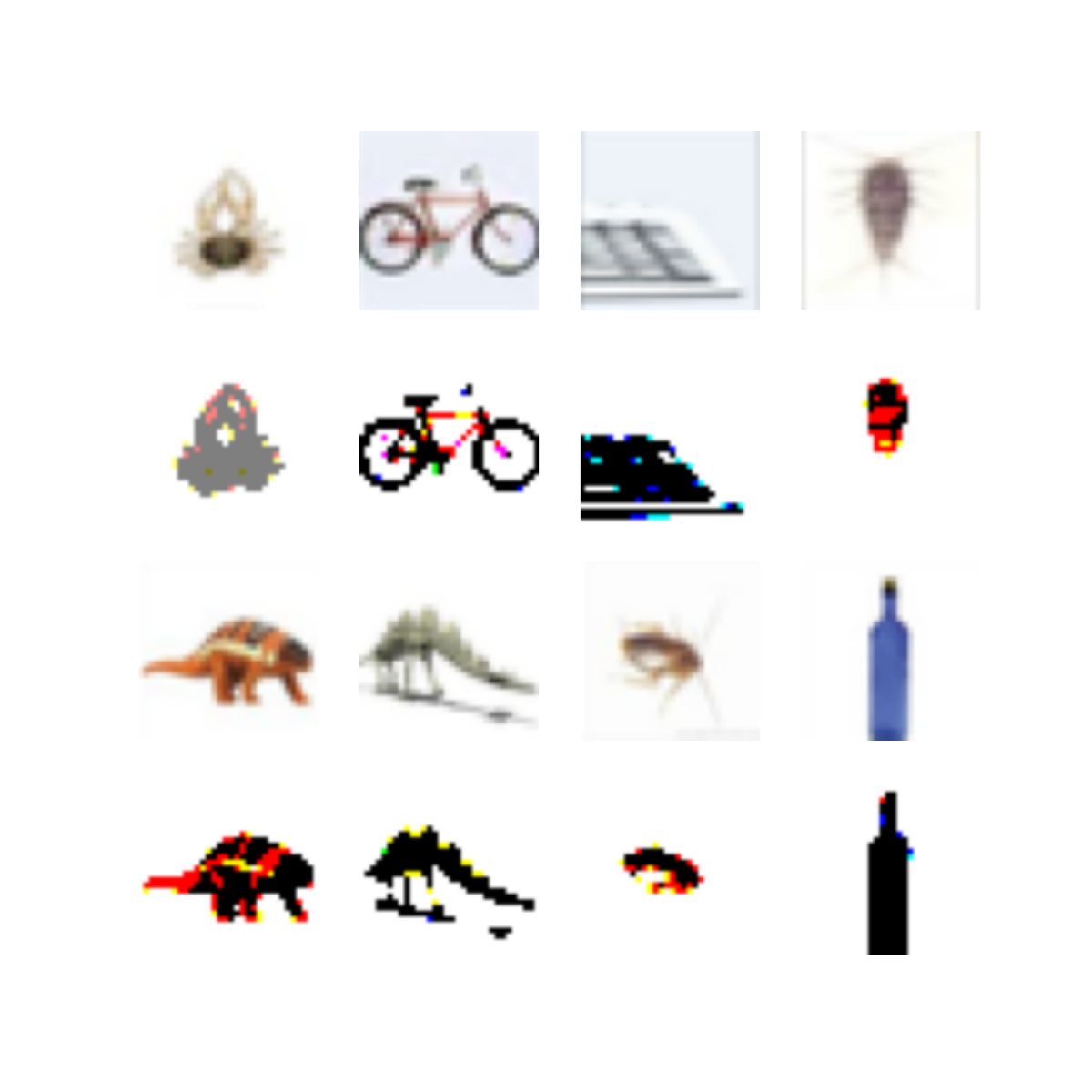}}
\end{center}
\vspace*{-3mm}
\caption{\label{fig:RtF_recon_clients} Top-8 SSIM reconstructed images for Robbing the Fed (RtF)~\cite{fowl2022robbing} for (a) 10, (b) 25, and (c) 50 clients in FedAVG. Clients train with 8 local iterations of mini-batch size 8 with $\alpha=1e-4$  on CIFAR-100. The 1st and 3rd rows are ground truth and the 2nd and 4th rows are the corresponding reconstructions. The leakage rate along with the quality of reconstructed images decreases with an increasing number of clients.}
\vspace*{-3mm}
\end{figure*}

Figure~\ref{fig:reconstruction_examples_cifar100} shows the reconstruction of a client for \name and RtF~\cite{fowl2022robbing} for the FedAVG attack on 100 clients. We do not use the bias for any reconstructions as discussed in Section~\ref{sec:reconstruction_equation}. The images reconstructed by \name are not affected by the aggregation of many clients and are very clearly identifiable. However, RtF is unable to reconstruct images correctly, with only 1 image having a non-zero computed weight gradient out of the FedAVG update. However, even after proper normalization, the image is visually incorrect due to precision errors. We find that a large majority of clients do not have even a single image with a non-zero gradient. 
Appendix Figure~\ref{fig:RtF_recon_clients} shows the top-8 SSIM reconstructed images across all clients in RtF for 10, 25, and 50 clients. As the number of clients increases, both the quantity and the quality of reconstructed images decrease.

We use the same FedAVG settings as above and additionally add 5 local epochs and compare the leakage rate for 100 clients. Table~\ref{tab:leakage_table} shows the leakage rate of \name and RtF + model inconsistency (MI)~\cite{pasquini2021eluding} on CIFAR-100, Tiny ImageNet, and MNIST. We do not compare to RtF alone, as it is unable to leak images in this setting. As discussed previously, increasing the FC layer size results in more precision problems and decreasing the size also does not help due to increased overlapping neuron activations. Using RtF with MI achieves the current SOTA for a large scale FedAVG secure aggregation attack and allows the attack to reach a single client. \name achieves a significantly higher leakage rate as it reaches all clients and scales to aggregation. Increasing the size of the FC layer will also allow the leakage rate to increase (for \name, as RtF runs into precision problems beyond a point as discussed in Section~\ref{sec:FedAVG_CSF}). 

\begin{figure}[!t]
\begin{center}
\includegraphics[width=0.9\columnwidth,trim={0 0 0 10mm},clip]{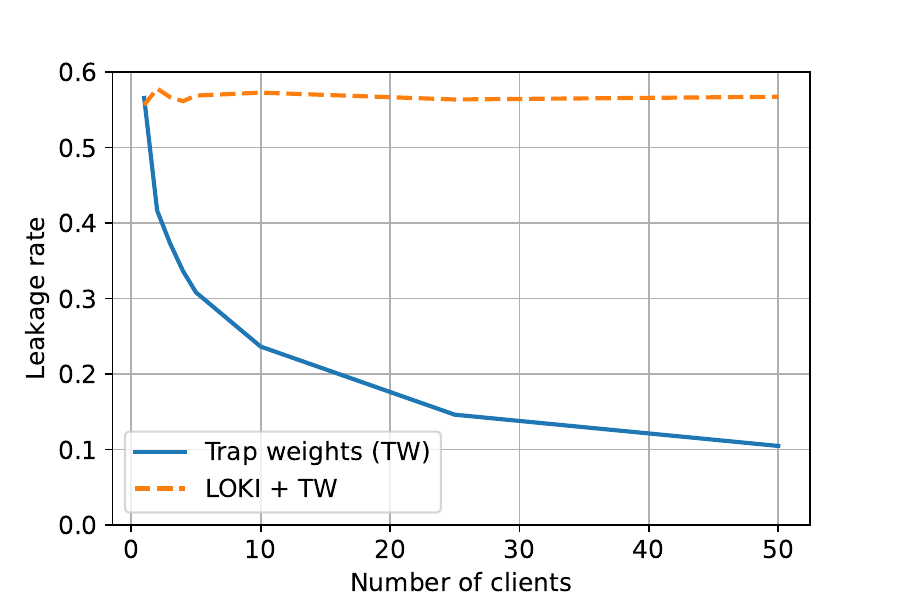}
\end{center}
\vspace*{-5mm}
\caption{\label{fig:wtc_leakage} Leakage rate on CIFAR-100 for the trap weights (TW) attack~\cite{boenisch2021curious} and \name + TW for a varying number of clients in FedSGD with a batch size of 64. Leakage rate given as an average over 10 runs. The TW attack leakage rate decreases as the number of clients increases, but using the convolutional attack fixes the scalability problem through split scaling.}
\vspace*{-5mm}
\end{figure}

\vspace{-2mm}
\subsection{FedSGD aggregation attack}
\vspace{-2mm}
We also show the applicability of \name to helping the trap weights (TW)~\cite{boenisch2021curious} attack in FedSGD aggregation. Figure~\ref{fig:wtc_leakage} shows the leakage rate of the TW attack and \name + TW for a varying number of clients between 1-50. Clients train with a batch size of 64 on CIFAR-100 and we use an FC layer size of $10\times$ the total number of images for the TW attack. For each setting of the number of clients, we use the TW scaling factor that achieves the highest leakage rate tested by 0.01 increments. For 1-50 clients, we find that scaling factors between 0.91-0.96 achieve the highest leakage rates, with lower values needed for higher numbers of clients. Despite tuning, the TW attack still suffers from a decreasing leakage rate as the number of clients increases. However, using the convolutional attack of \name in addition to TW allows the scaling to split between the batch size and number of clients and maintains a constant leakage rate regardless of the number of clients.

\begin{figure}[t!]
\begin{center}     
\subfigure[Leakage rate]{\label{fig:dataset_size_leakage_rate}\includegraphics[width=0.5\columnwidth,trim={3mm 0mm 3mm 10mm},clip]{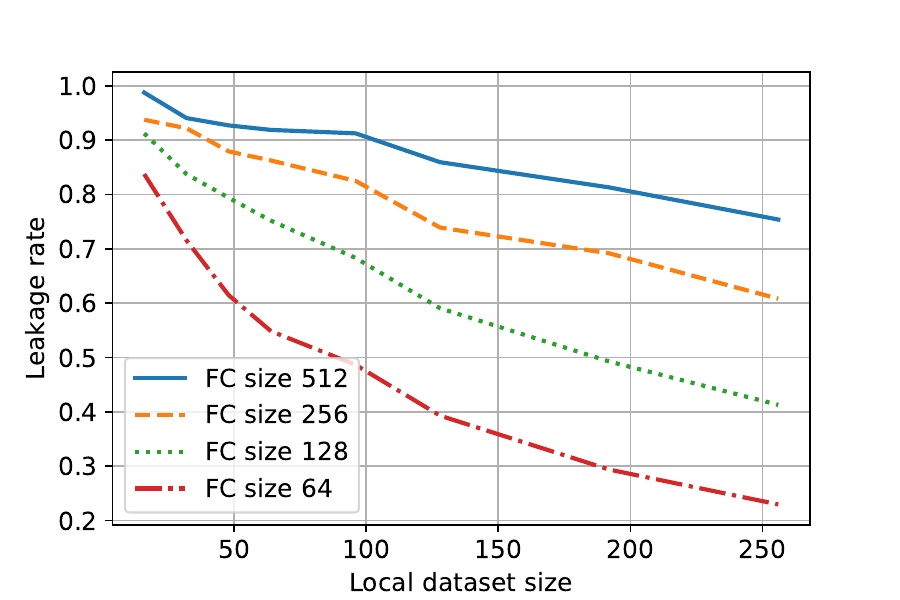}}
\hspace{-4 mm}
\subfigure[Number of leaked images]
{\label{fig:dataset_size_images}\includegraphics[width=0.5\columnwidth,trim={3mm 0mm 3mm 10mm},clip]{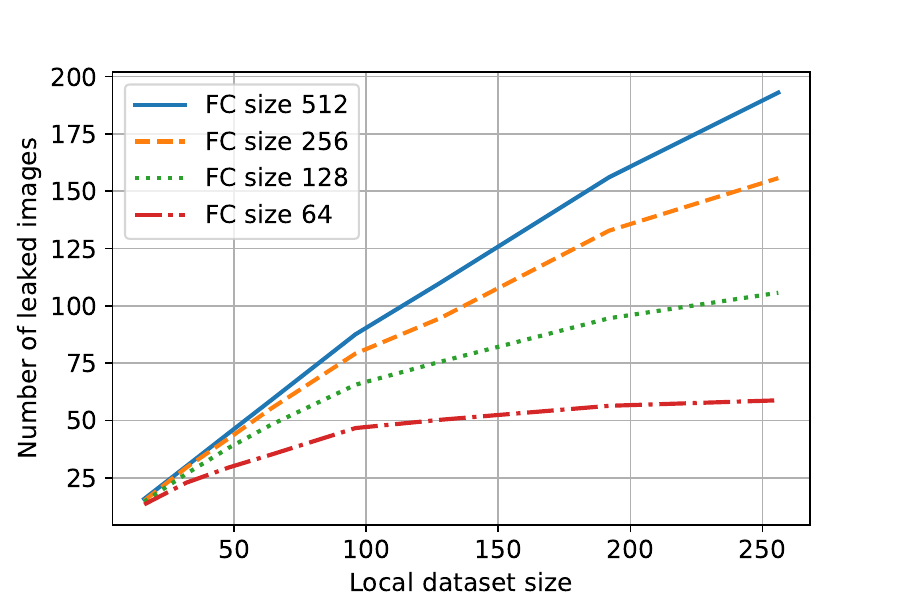}}
\end{center}
\vspace*{-4mm}
\caption{For \name, (a) leakage rate and (b) number of leaked images as a function of the local dataset size and FC layer size averaged over 10 clients. While the overall leakage rate decreases with a larger local dataset size, the total number of leaked images continues to increase.}
\end{figure}

\begin{figure}[t!]
\begin{center}
\includegraphics[width=0.9\columnwidth,trim={0mm 0mm 0mm 11mm},clip]{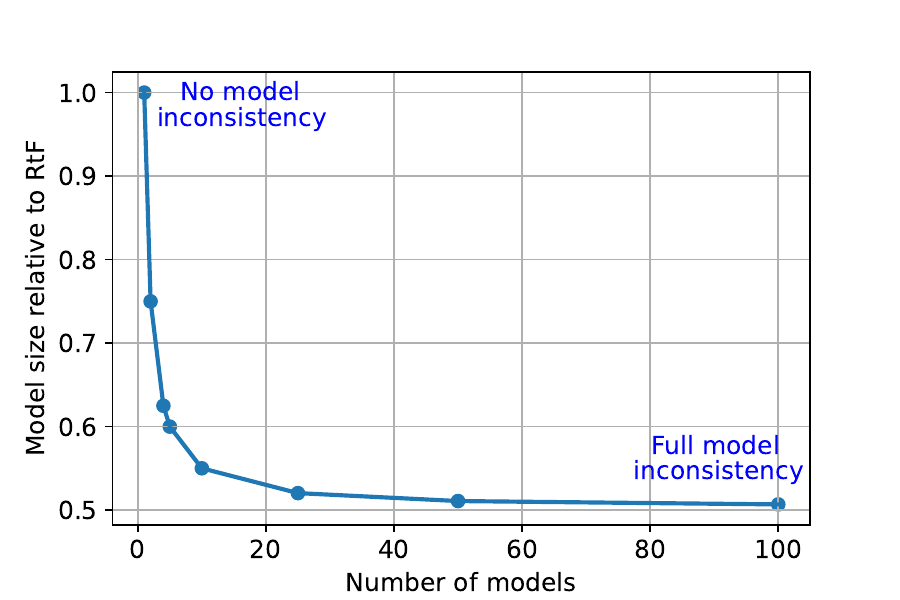}
\end{center}
\vspace*{-5mm}
\caption{\label{fig:model-inconsistency-size} Model size of \name relative to the size added by Robbing the Fed (RtF) for a 100 client attack. The number of clients sharing each separate model is equal across all models. \name's size is $50.69\%-100\%$ of RtF's.}
\vspace*{-5mm}
\end{figure}

\vspace{-1mm}
\subsection{Local dataset size and FC layer size}
\vspace{-2mm}
The leakage rate is affected by the local dataset size and the FC layer size. In this section we experiment with how the changes to both affect overall leakage rate. We fix the local mini-batch size to 8 and vary the number of local iterations as the dataset size increases. We use $\alpha=1e-4$ and $\textit{CSF}=500$ for all settings and train on CIFAR-100. For each local dataset size, we compute the leakage rate with an FC layer size of 64, 128, 256, and 512 as an average over 10 clients. Figure~\ref{fig:dataset_size_leakage_rate} shows the average leakage rate for each setting as the local dataset size increases from 32-256.

Figure~\ref{fig:dataset_size_images} shows the average number of leaked images per client with the varying local dataset size. While the overall leakage rate decreases as the local dataset size increases, the total number of leaked images continues to increase. The local iteration parameter changes prevent multiple images between separate local iterations from activating the same neuron. As a result, even as the ratio of images to neurons increases, the number of leaked images can also continue to increase without having issues with multiple activations on the same neuron preventing reconstruction, up until the point where nearly each neuron leaks a separate image. Appendix Figure~\ref{fig:dataset_size_local_iterations} shows the effect of the number of local iterations on the leakage rate.

\vspace{-1mm}
\subsection{Convolutional scaling factor}
\label{sec:csf_experiments}
\vspace{-2mm}
While the convolutional scaling factor (\textit{CSF}) allows us to have attacks without model inconsistency, there is a trade-off in model size. Figure~\ref{fig:model-inconsistency-size} shows the model size for varying levels of model inconsistency (number of clients sharing the same model) relative to the size added by RtF when they both achieve the same leakage rate in FedSGD attacking 100 clients in aggregation (FedSGD instead of FedAVG, since RtF cannot scale in FedAVG due to precision). This happens when the effective number of bins, $\textit{number of identity mapping sets}\times\textit{FC layer size}$, is the same as the total FC layer size in RtF. With full model inconsistency, \name adds only $50.69\%$ of the size of the RtF attack. With no model inconsistency, \name is the same size at $100.00\%$. \name maintains the same leakage rate for all points. Recall that RtF has no model inconsistency. 

\begin{figure}[t!]
\begin{center}     
\subfigure[Leakage rate and PSNR]{\label{fig:csf-leak-psnr}\includegraphics[width=0.51\columnwidth,trim={0 4mm 0 3mm},clip]{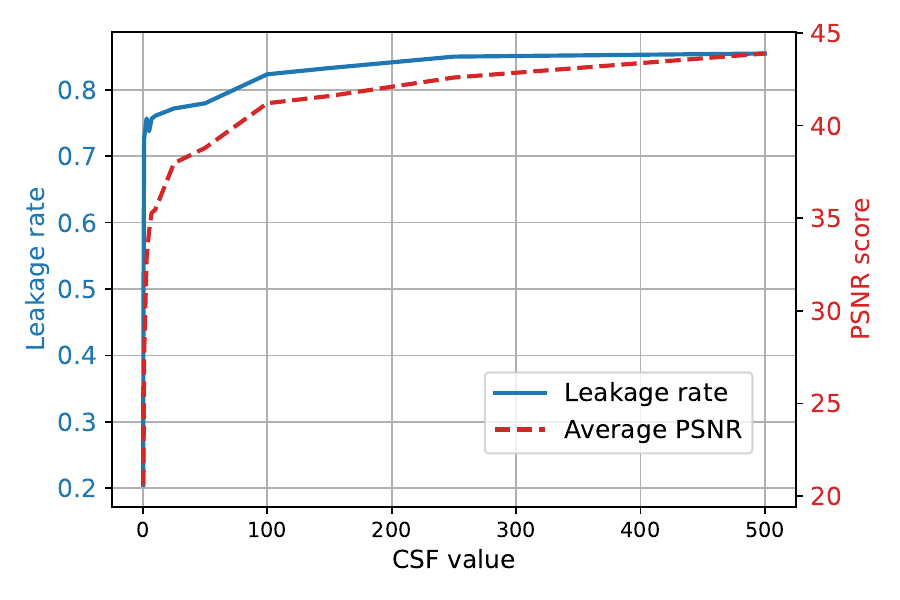}}
\hspace{-4mm}
\subfigure[SSIM and LPIPS]{\label{fig:csf-ssim-lpips}\includegraphics[width=0.51\columnwidth,trim={0 4mm 0 3mm},clip]{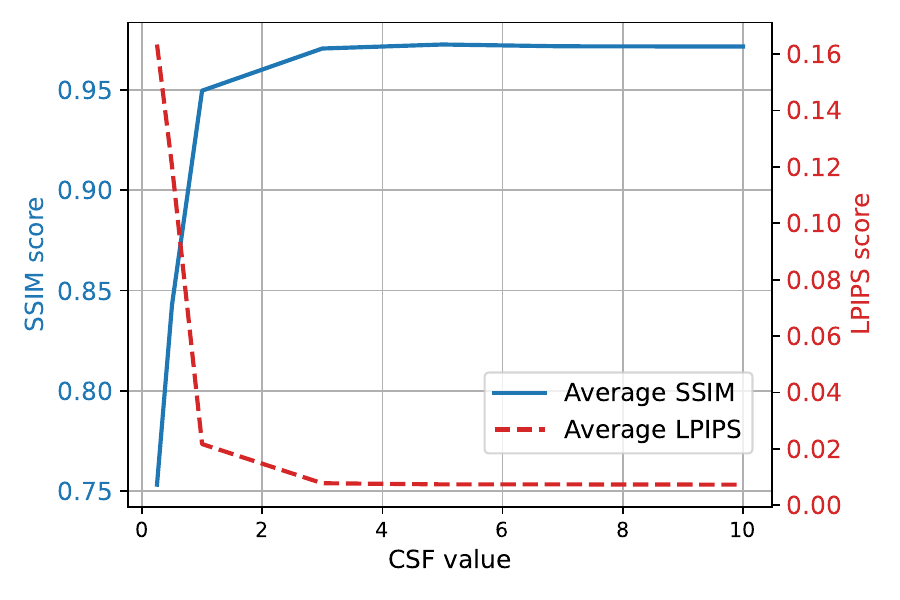}}
\end{center}
\vspace*{-4mm}
\caption{\label{fig:csf_metrics} For \name (a) leakage rate and PSNR$\uparrow$, (b) SSIM$\uparrow$ and LPIPS$\downarrow$ as a function of the \textit{CSF}. The leakage rate and PSNR increase over a larger range of \textit{CSF} values while the SSIM and LPIPS scores stop improving much quicker.}
\vspace*{-5mm}
\end{figure}

The other purpose of the CSF is to mitigate changes in the identity mapping kernels and prevent images from activating the same neurons. Figure~\ref{fig:csf_metrics} show the leakage rate and average PSNR$\uparrow$, SSIM$\uparrow$, and LPIPS$\downarrow$ scores of reconstructed images when using various \textit{CSF} values (we report the average of reconstructed images without activation overlap, even when the SSIM is below 0.5 in order to properly show the impact of the \textit{CSF} on the metrics).
We average the values over 10 clients in FedAVG aggregation training on CIFAR-100 using 8 local iterations of mini-batch size 8 and $\alpha=1e-4$. With very small \textit{CSF} values, the leakage rate is much lower because the convolutional kernel parameter changes prevent images from being pushed through correctly. Similarly, we see lower metric scores for reconstructed images. High \textit{CSF} values result in larger changes to the FC layer weights and a higher leakage rate by preventing images from activating the same neurons. Due to smaller changes in the identity mapping sets, the reconstruction metrics also improve. The PSNR score improves over a large range of CSF values (continuing beyond $\textit{CSF}=500$) while the SSIM and LPIPS score improvement stops much quicker. 
For example, the attack achieves a $72.59\%$ leakage rate when $\textit{CSF}=1$ and increases until $\textit{CSF}=500$ where it peaks at $85.47\%$ and no longer increases with larger \textit{CSF} values.

\vspace{-1mm}
\subsection{Non-IID federated learning}
\vspace{-1mm}
Previous experiments worked under the setting that client data was IID and the biases were initialized the same for everyone. In these experiments, the clients contain non-IID distributions. We use OrganAMNIST instead of the CIFAR dataset, as it has a less uniform average pixel intensity distribution which is shown in Appendix Figure~\ref{fig:distribution-datasets}. For OrganAMNIST, label-based separation also results in individual distributions with larger differences in mean and SD. For this separation, each client has data from a single class. Since OrganAMNIST only has 11 classes, several clients share data from the same label. However, this is not known to the server so the distributions of clients are learned independently. We use a dataset agnostic bias initialization for the first round, where all clients have initial biases with mean -0.5 and SD 0.25 following a normal distribution. This value is adjusted after each training round separately for each client based on observing the activated biases.

\begin{figure}[t!]
\begin{center}
\includegraphics[width=0.9\columnwidth,trim={0 0 0 10mm},clip]{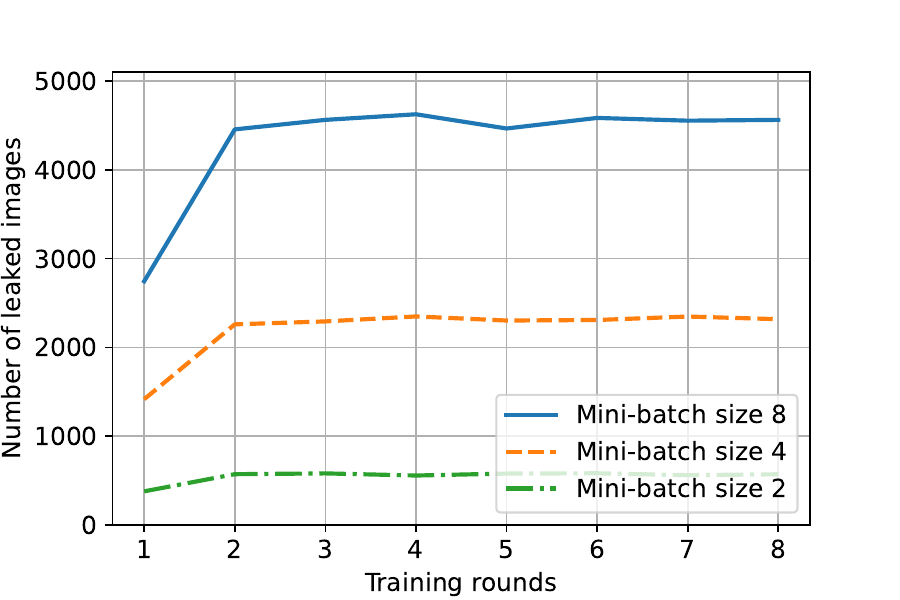}
\end{center}
\vspace*{-5mm}
\caption{\label{fig:organ-leakage-results} [Effect of non-IID clients] Number of images leaked for various local mini-batch sizes training on the OrganAMNIST dataset with $100$ non-IID clients over several training rounds. The first training iteration starts with a dataset agnostic bias initialization and subsequent training rounds improve through observing activations. The attacks improve significantly after a single round.}
\vspace*{-3mm}
\end{figure}

We use $100$ clients for each experiment each training over 8 local iterations, but we vary the local mini-batch size as $8$, $4$, and $2$ (total of $64$, $32$, and $16$ images). The FC layer size is setup equal to $4\times$ the total number of images and we use $\textit{CSF}=100$. Figure~\ref{fig:organ-leakage-results} shows the number of images leaked over several training rounds for the different local mini-batch sizes. The total number of images for each local mini-batch size is $6400$, $3200$, and $1600$ respectively. After just the first training round of observing neuron activations, a jump in leaked images occurs for all mini-batch sizes. An increase of $62.32\%$, $59.53\%$, and $50.92\%$ leaked images is observed for local mini-batch sizes $8$, $4$, and $2$ respectively. After the initial jump, the total leakage fluctuates slightly between rounds due to randomness of the client mini-batch. 

\begin{figure}[t!]
\begin{center}
\includegraphics[width=0.9\columnwidth,trim={0 0 0 2mm},clip]{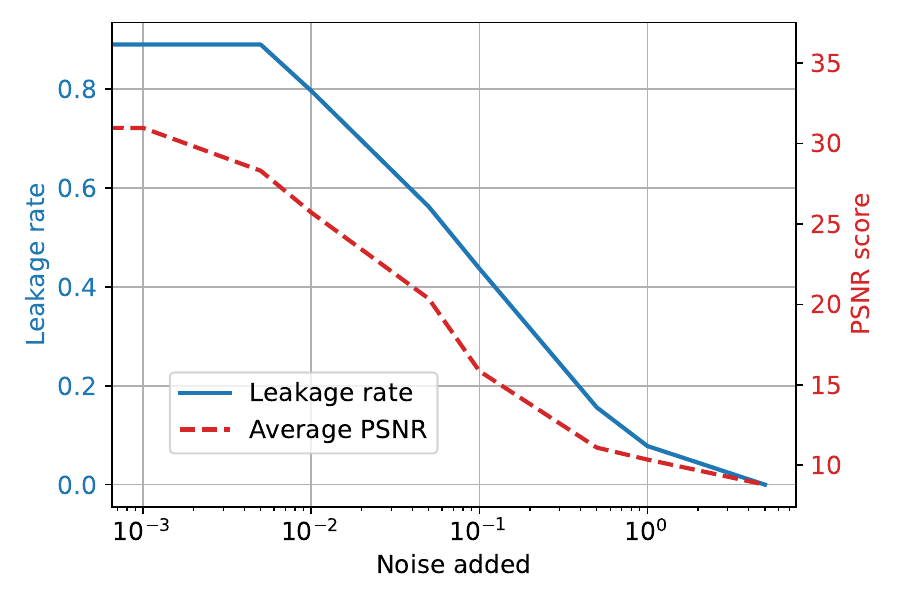}
\end{center}
\vspace*{-6mm}
\caption{\label{fig:differential_privacy} Leakage rate and average PSNR for client reconstruction in CIFAR-100 under $\sigma$ Gaussian noise between 1e-3 and 5. Images with an SSIM $> 0.5$ are counted as leaked. At $\sigma=5$, no images are leaked.}
\vspace*{-5mm}
\end{figure}

\vspace{-1mm}
\subsection{Adding noise}
\begin{figure*}[t!]
\begin{center}     
\vspace{-3mm}
\subfigure[$\sigma=1e-3$]{\label{fig:dp_1e-3}\includegraphics[width=0.65\columnwidth,trim={22mm 22mm 22mm 22mm},clip]{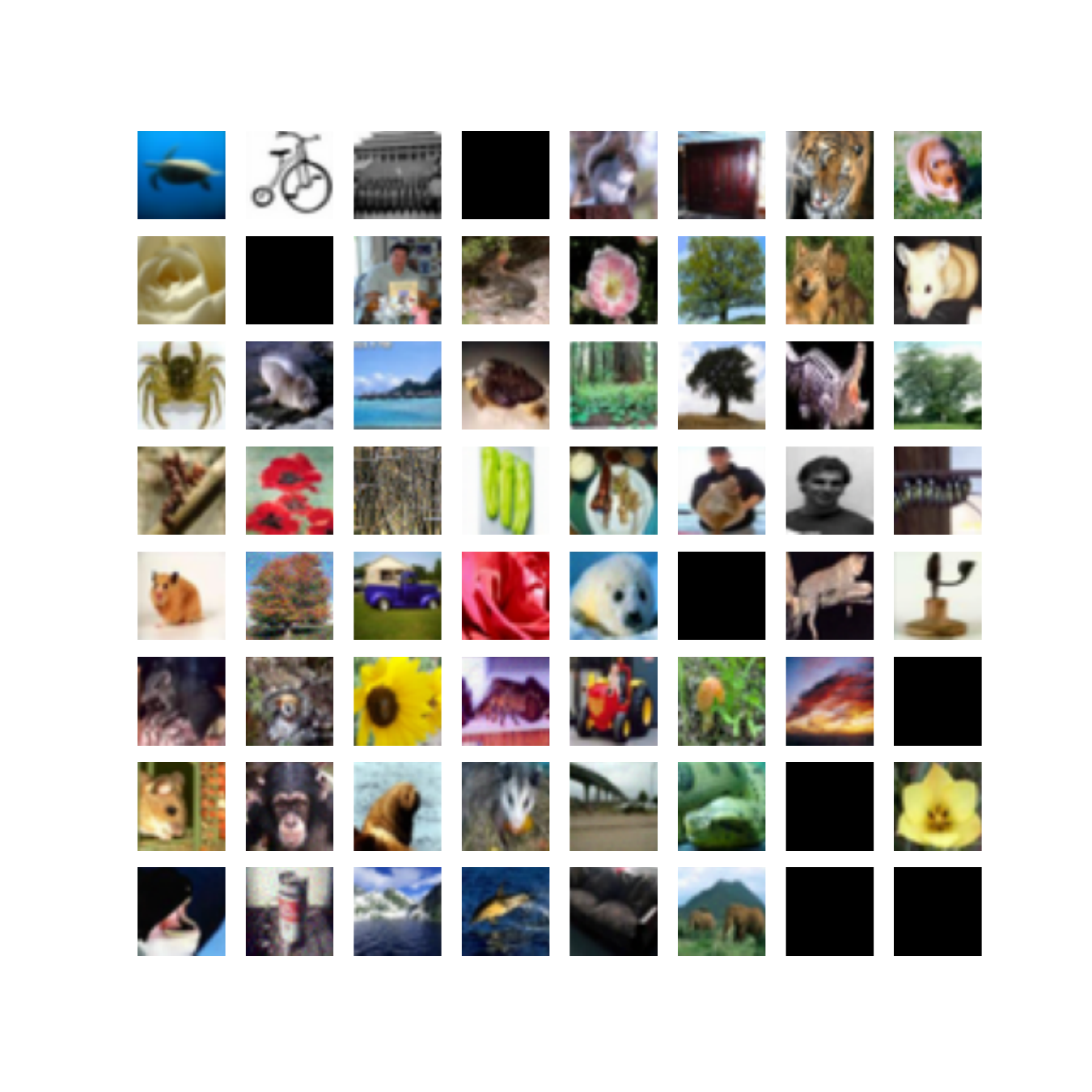}}\hspace{5mm}%
\subfigure[$\sigma=1e-1$]{\label{fig:dp_1e-1}\includegraphics[width=0.65\columnwidth,trim={22mm 22mm 22mm 22mm},clip]{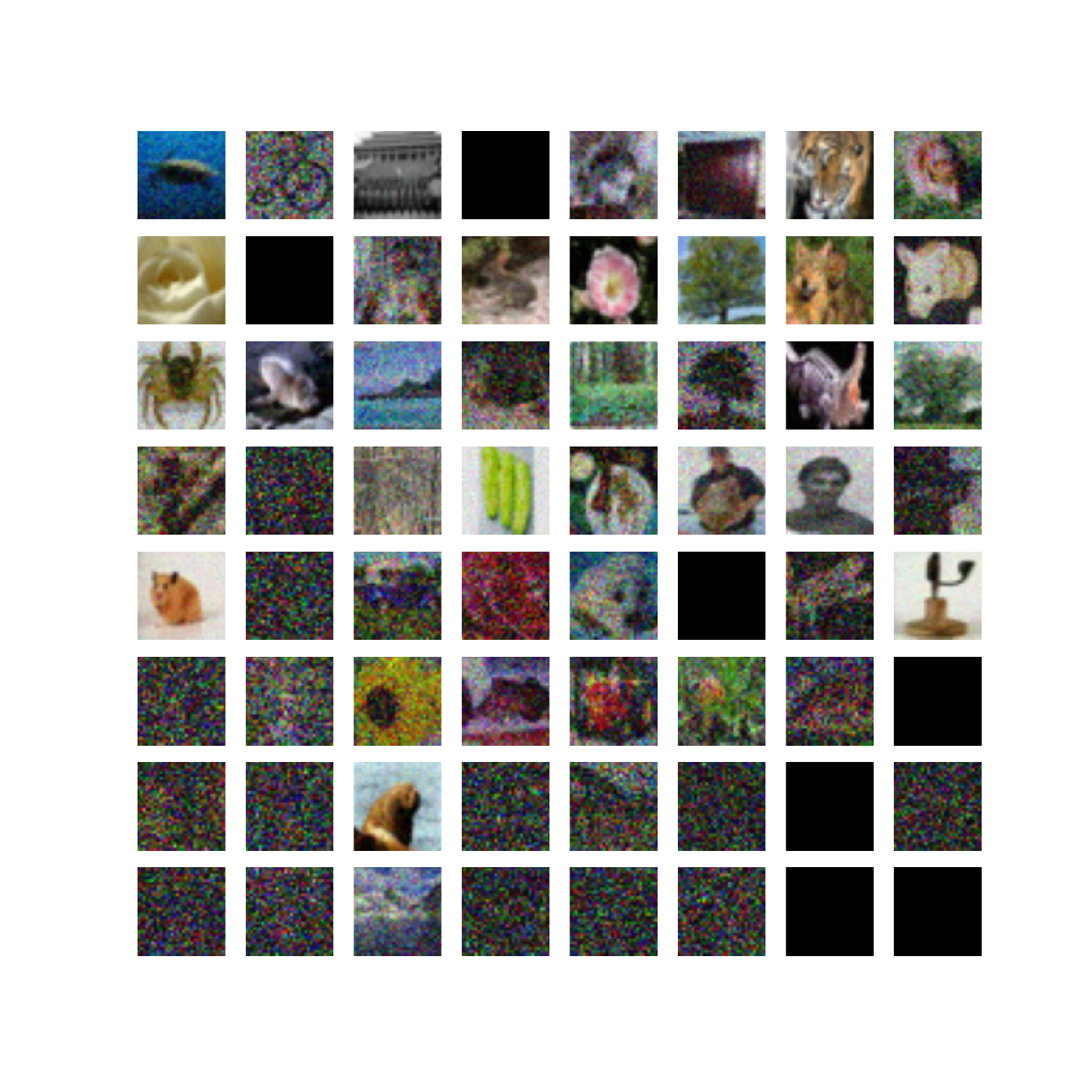}}\hspace{5mm}%
\subfigure[$\sigma=5$]{\label{fig:dp_5}\includegraphics[width=0.65\columnwidth,trim={22mm 22mm 22mm 22mm},clip]{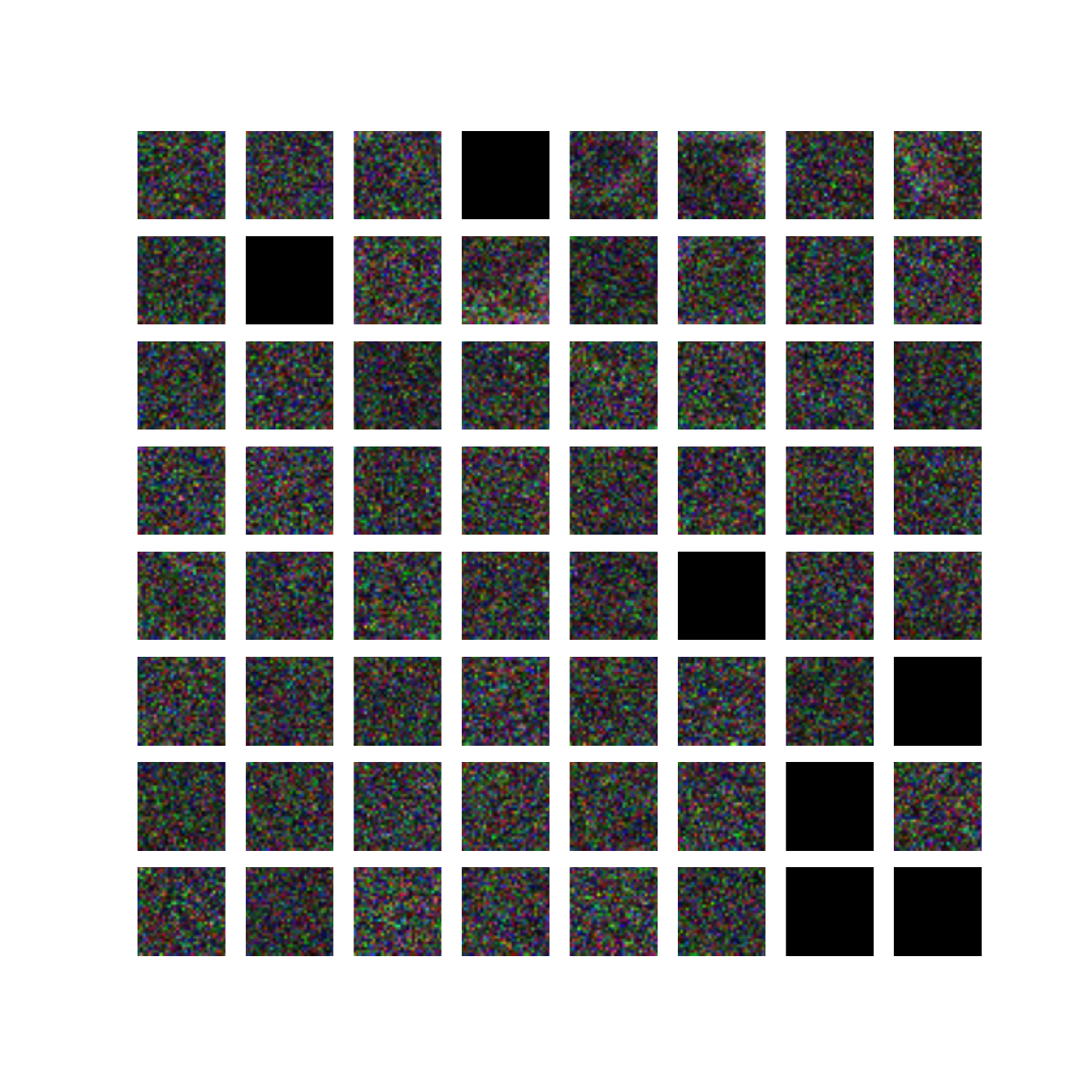}}
\end{center}
\vspace*{-3mm}
\caption{\label{fig:dp_reconstructions} Reconstruction examples for a client with varying $\sigma$ noise added to the update. Client training with $\alpha=1e-1$, 4 local iterations of mini-batch size 16 and $\textit{CSF}=500$. When (a) $\sigma=1e-3$ the attack has the maximum leakage rate and the images are clearly identifiable. When (b) $\sigma=1e-1$ the average SSIM is lower, but some images are visually identifiable. With (c) $\sigma=5$ all images are unidentifiable.}
\vspace*{-5mm}
\end{figure*}

\vspace{-1mm}
Applying noise locally is used as a common form of defense against privacy attacks. We explore the effects of adding different levels of Gaussian noise to the attack on a single client. We quantify the effects of noise through the average reconstructed image PSNR score and the leakage rate (reconstructions with SSIM $> 0.5$). We use $\textit{CSF}=500$ in order to achieve the maximum default leakage rate, $\alpha=1e-1$, and train with 4 local iterations of mini-batch size 16. Compared to FedSGD, the learning rate $\alpha$ in FedAVG directly impacts the size of the computed gradient magnitude. Figure~\ref{fig:differential_privacy} shows the PSNR and leakage rate for various standard deviation $\sigma$ amounts of noise between $1e-3$ and 5 added to the update trained on CIFAR-100. Only when the amount of noise added is $\sigma=5$, is all image leakage prevented. However, while adding $\sigma=5$ noise to the model can prevent leakage, it will also destroy a model's training accuracy. Appendix Figure~\ref{fig:dp_reconstructions} shows reconstruction results with added noise.

Increasing the number of clients in secure aggregation also increases the amount of noise added to the reconstructions. The noise in the aggregation works as adding multiple gaussian distributions together. The mean of the cumulative gaussian is the sum of the means of the individual distributions (in this case it is 0) and the variance is the sum of individual variances. In the above figure, the attacks achieves 0 leakage with added noise of STD $\sigma=5$. If each client uses $\sigma=5e-2$, having $N=\frac{(5)^2}{(5e-2)^2} = 10,000$ clients in aggregation will achieve the same amount of noise as a single client adding $\sigma=5$. 
\label{sec:experiments}

\vspace{-1mm}
\section{Defenses and mitigation}
\vspace{-2mm}
As a major outcome of our work, we find that FedAVG using secure aggregation for defense does {\em not} prevent a malicious server from recovering large amounts of private user data regardless of the number of clients participating. The same is true for secure shuffling, discussed in the context of federated learning by~\cite{kairouz2021advances}. Secure shuffling prevents a server from receiving any additional information (outside of the update itself) which would allow identification of which client sent each individual message. However, secure shuffling cannot nullify our attack as the update itself carries the fingerprint of the client. While reconstructing the image, the server can identify the client it came from. 

One possibility of defense is through identification of a modified model architecture or the parameters. While \name uses a certain order of layers (convolutional layer followed by two FC layers), this module can be difficult to identify as the layers used can be further in the model architecture. Models can use multiple convolutional layers followed by FC layers. In this case, \name can use the final convolutional layer and subsequent FC layers as the leakage module. Figure~\ref{fig:imagenet-downsample} in the appendix shows reconstructions after max-pooling. While the reconstructions lose resolution, the content is still clear. \name is similar to Robbing the Fed~\cite{fowl2022robbing} in stealthiness. Although it requires an additional convolutional layer, identification is just as difficult since it is easy to place this in multiple locations of the benign architecture of models such as ResNet or VGG by having prior layers act to push the input forward. A client could also attempt to identify malicious tampering of parameters prior to training on local data and sending an update. However, this could also be difficult due to the server's ability to mask the weight gradient in different ways to have the non-identity mapping sets output zero. Furthermore, the no model inconsistency attack does not use zero parameters. There is also still a fundamental problem with identification in that clients have limited computational abilities in the setting of cross-device FL, which may preclude such verification. They typically only follow a standard federated learning protocol given to them that consists of training the model and sending updates back to the server. 

Model inconsistency defense has also been proposed in~\cite{pasquini2021eluding}. Having clients utilize a pseudorandom generator for SA using a condition that the received models are the same will prevent the server from being able to unencrypt the SA updates if it sends different models to each client. This method incurs no additional communication overhead and is applicable to SA algorithms such as~\cite{bonawitz2017practical,secagg_bell2020secure}. However, the practical scenario of asynchronous FL also uses SA~\cite{so2021lightsecagg,nguyen2022federated} but inherently has model inconsistency due to client staleness. Methods for on-device efficient training also send clients models with different architectures~\cite{diao2021heterofl}. Furthermore, the server does not absolutely require model inconsistency for \name. By choosing to not use model inconsistency, the leakage rate of the attack is slightly lower, but the server can increase the stealth of the attack with no impact on the reconstruction quality or scalability (Figure~\ref{fig:scalability_rtf}).

One standard mitigation strategy in federated learning would be to use differential privacy~\cite{dwork2014algorithmic,jayaraman2019evaluating,wei2020federated} to add noise to updates prior to sending them to the server. While this method can be effective in preventing a server from fully reconstructing private data, it comes with the hefty downside in a decrease of model performance, especially with large vision models. Particularly, the large dimensionality of many vision models causes the impact of noise to be greater when achieving the same privacy guarantees~\cite{zhang2022understanding}. The modification of the model due to the attack does not need to occur during every step of the training process. Without knowledge of when the attack will occur, differential privacy must then be applied during every step of the training process, thus seriously impacting accuracy.
\label{sec:defenses}

\vspace{-1mm}
\section{Conclusions}
\vspace{-2mm}
In this paper, we have demonstrated how to break the privacy of secure aggregation in FedAVG federated learning through a malicious server that sends customized models to clients. Our key design idea is to send customized convolutional kernels to each client, an identity mapping set, that separates the weight gradients from the clients despite the use of secure aggregation. The server then uses these weight gradients to reconstruct the original data points. Using our proposed convolutional scaling factor, the attack can avoid model inconsistency and achieve a higher leakage rate in FedAVG than FedSGD attacks. We are the first to achieve a privacy attack in FL with FedAVG that scales well with the size of the local dataset and the number of clients. For us to handle an increasing local dataset size, the fully connected layer size increases linearly. To handle an increasing number of clients, the size of the convolutional layer increases linearly and so the total number of parameters grows linearly, while the number of non-zero parameters stays constant. We achieve high reconstruction quality and a leakage rate between 76-86\% for CIFAR-100, Tiny ImageNet, and  MNIST with 100 clients, while that of the state-of-the-art is less than 1\%.
\label{sec:conclusions}

\smallbreak
\noindent \textbf{Acknowledgements.} This work was supported by Army Research Lab under Contract No. W911NF-2020-221, National Science Foundation CNS-2038986, Defense Advanced Research Projects Agency (DARPA) under Contract No. HR001120C0156, ARO award W911NF1810400, and ONR Award No. N00014-16-1-2189. Any opinions, findings, and conclusions or recommendations expressed in this material are those of the authors and do not necessarily reflect the views of the sponsors.

\bibliographystyle{IEEEtran}
\bibliography{references}

\begin{thebibliography}{10}
\providecommand{\url}[1]{#1}
\csname url@samestyle\endcsname
\providecommand{\newblock}{\relax}
\providecommand{\bibinfo}[2]{#2}
\providecommand{\BIBentrySTDinterwordspacing}{\spaceskip=0pt\relax}
\providecommand{\BIBentryALTinterwordstretchfactor}{4}
\providecommand{\BIBentryALTinterwordspacing}{\spaceskip=\fontdimen2\font plus
\BIBentryALTinterwordstretchfactor\fontdimen3\font minus \fontdimen4\font\relax}
\providecommand{\BIBforeignlanguage}[2]{{%
\expandafter\ifx\csname l@#1\endcsname\relax
\typeout{** WARNING: IEEEtran.bst: No hyphenation pattern has been}%
\typeout{** loaded for the language `#1'. Using the pattern for}%
\typeout{** the default language instead.}%
\else
\language=\csname l@#1\endcsname
\fi
#2}}
\providecommand{\BIBdecl}{\relax}
\BIBdecl

\bibitem{mcmahan2017communication}
B.~McMahan, E.~Moore, D.~Ramage, S.~Hampson, and B.~A. y~Arcas, ``Communication-efficient learning of deep networks from decentralized data,'' in \emph{Artificial intelligence and statistics}.\hskip 1em plus 0.5em minus 0.4em\relax PMLR, 2017, pp. 1273--1282.

\bibitem{melis2019exploiting}
L.~Melis, C.~Song, E.~De~Cristofaro, and V.~Shmatikov, ``Exploiting unintended feature leakage in collaborative learning,'' in \emph{2019 IEEE symposium on security and privacy (SP)}.\hskip 1em plus 0.5em minus 0.4em\relax IEEE, 2019, pp. 691--706.

\bibitem{luo2021feature}
X.~Luo, Y.~Wu, X.~Xiao, and B.~C. Ooi, ``Feature inference attack on model predictions in vertical federated learning,'' in \emph{2021 IEEE 37th International Conference on Data Engineering (ICDE)}.\hskip 1em plus 0.5em minus 0.4em\relax IEEE, 2021, pp. 181--192.

\bibitem{shokri2017membership}
R.~Shokri, M.~Stronati, C.~Song, and V.~Shmatikov, ``Membership inference attacks against machine learning models,'' in \emph{2017 IEEE symposium on security and privacy (SP)}.\hskip 1em plus 0.5em minus 0.4em\relax IEEE, 2017, pp. 3--18.

\bibitem{choquette2021label}
C.~A. Choquette-Choo, F.~Tramer, N.~Carlini, and N.~Papernot, ``Label-only membership inference attacks,'' in \emph{International conference on machine learning}.\hskip 1em plus 0.5em minus 0.4em\relax PMLR, 2021, pp. 1964--1974.

\bibitem{nasr2019comprehensive}
M.~Nasr, R.~Shokri, and A.~Houmansadr, ``Comprehensive privacy analysis of deep learning: Passive and active white-box inference attacks against centralized and federated learning,'' in \emph{2019 IEEE symposium on security and privacy (SP)}.\hskip 1em plus 0.5em minus 0.4em\relax IEEE, 2019, pp. 739--753.

\bibitem{hitaj2017deep}
B.~Hitaj, G.~Ateniese, and F.~Perez-Cruz, ``Deep models under the gan: information leakage from collaborative deep learning,'' in \emph{Proceedings of the 2017 ACM SIGSAC conference on computer and communications security}, 2017, pp. 603--618.

\bibitem{wang2019beyond}
Z.~Wang, M.~Song, Z.~Zhang, Y.~Song, Q.~Wang, and H.~Qi, ``Beyond inferring class representatives: User-level privacy leakage from federated learning,'' in \emph{IEEE INFOCOM 2019-IEEE Conference on Computer Communications}.\hskip 1em plus 0.5em minus 0.4em\relax IEEE, 2019, pp. 2512--2520.

\bibitem{zhu2019deep}
L.~Zhu, Z.~Liu, and S.~Han, ``Deep leakage from gradients,'' \emph{Advances in neural information processing systems}, vol.~32, 2019.

\bibitem{zhao2020idlg}
B.~Zhao, K.~R. Mopuri, and H.~Bilen, ``idlg: Improved deep leakage from gradients,'' \emph{arXiv preprint arXiv:2001.02610}, 2020.

\bibitem{geiping2020inverting}
J.~Geiping, H.~Bauermeister, H.~Dr{\"o}ge, and M.~Moeller, ``Inverting gradients-how easy is it to break privacy in federated learning?'' \emph{Advances in Neural Information Processing Systems}, vol.~33, pp. 16\,937--16\,947, 2020.

\bibitem{yin2021see}
H.~Yin, A.~Mallya, A.~Vahdat, J.~M. Alvarez, J.~Kautz, and P.~Molchanov, ``See through gradients: Image batch recovery via gradinversion,'' in \emph{Proceedings of the IEEE/CVF Conference on Computer Vision and Pattern Recognition}, 2021, pp. 16\,337--16\,346.

\bibitem{dimitrov2022data}
D.~I. Dimitrov, M.~Balunovic, N.~Konstantinov, and M.~Vechev, ``Data leakage in federated averaging,'' \emph{Transactions on Machine Learning Research}, 2022.

\bibitem{bonawitz2017practical}
K.~Bonawitz, V.~Ivanov, B.~Kreuter, A.~Marcedone, H.~B. McMahan, S.~Patel, D.~Ramage, A.~Segal, and K.~Seth, ``Practical secure aggregation for privacy-preserving machine learning,'' in \emph{proceedings of the 2017 ACM SIGSAC Conference on Computer and Communications Security}, 2017, pp. 1175--1191.

\bibitem{fereidooni2021safelearn}
H.~Fereidooni, S.~Marchal, M.~Miettinen, A.~Mirhoseini, H.~M{\"o}llering, T.~D. Nguyen, P.~Rieger, A.-R. Sadeghi, T.~Schneider, H.~Yalame \emph{et~al.}, ``Safelearn: secure aggregation for private federated learning,'' in \emph{2021 IEEE Security and Privacy Workshops (SPW)}.\hskip 1em plus 0.5em minus 0.4em\relax IEEE, 2021, pp. 56--62.

\bibitem{so2021lightsecagg}
J.~So, C.~J. Nolet, C.-S. Yang, S.~Li, Q.~Yu, R.~E~Ali, B.~Guler, and S.~Avestimehr, ``Lightsecagg: a lightweight and versatile design for secure aggregation in federated learning,'' \emph{Proceedings of Machine Learning and Systems}, vol.~4, pp. 694--720, 2022.

\bibitem{fowl2022robbing}
L.~H. Fowl, J.~Geiping, W.~Czaja, M.~Goldblum, and T.~Goldstein, ``Robbing the fed: Directly obtaining private data in federated learning with modified models,'' in \emph{International Conference on Learning Representations}, 2022.

\bibitem{boenisch2021curious}
F.~Boenisch, A.~Dziedzic, R.~Schuster, A.~S. Shamsabadi, I.~Shumailov, and N.~Papernot, ``When the curious abandon honesty: Federated learning is not private,'' \emph{8th IEEE European Symposium on Security and Privacy (IEEE Euro S\&P)}, 2023.

\bibitem{wen2022fishing}
Y.~Wen, J.~Geiping, L.~Fowl, M.~Goldblum, and T.~Goldstein, ``Fishing for user data in large-batch federated learning via gradient magnification,'' \emph{International Conference on Machine Learning}, 2022.

\bibitem{pasquini2021eluding}
D.~Pasquini, D.~Francati, and G.~Ateniese, ``Eluding secure aggregation in federated learning via model inconsistency,'' in \emph{Proceedings of the 2022 ACM SIGSAC Conference on Computer and Communications Security}, 2022, pp. 2429--2443.

\bibitem{kariyappa2022cocktail}
S.~Kariyappa, C.~Guo, K.~Maeng, W.~Xiong, G.~E. Suh, M.~K. Qureshi, and H.-H.~S. Lee, ``Cocktail party attack: Breaking aggregation-based privacy in federated learning using independent component analysis,'' in \emph{International Conference on Machine Learning}.\hskip 1em plus 0.5em minus 0.4em\relax PMLR, 2023, pp. 15\,884--15\,899.

\bibitem{lam2021gradient}
M.~Lam, G.-Y. Wei, D.~Brooks, V.~J. Reddi, and M.~Mitzenmacher, ``Gradient disaggregation: Breaking privacy in federated learning by reconstructing the user participant matrix,'' in \emph{International Conference on Machine Learning}.\hskip 1em plus 0.5em minus 0.4em\relax PMLR, 2021, pp. 5959--5968.

\bibitem{cho2020client}
Y.~J. Cho, J.~Wang, and G.~Joshi, ``Client selection in federated learning: Convergence analysis and power-of-choice selection strategies,'' \emph{arXiv preprint arXiv:2010.01243}, 2020.

\bibitem{chen2020optimal}
W.~Chen, S.~Horv{\'a}th, and P.~Richt{\'a}rik, ``Optimal client sampling for federated learning,'' \emph{Transactions on Machine Learning Research}, 2022.

\bibitem{pejo2020quality}
B.~Pej{\'o} and G.~Bicz{\'o}k, ``Quality inference in federated learning with secure aggregation,'' \emph{arXiv preprint arXiv:2007.06236}, 2020.

\bibitem{secagg_so2021securing}
J.~So, R.~E. Ali, B.~Guler, J.~Jiao, and S.~Avestimehr, ``Securing secure aggregation: Mitigating multi-round privacy leakage in federated learning,'' \emph{arXiv preprint arXiv:2106.03328}, 2021.

\bibitem{phong2017privacy}
L.~T. Phong, Y.~Aono, T.~Hayashi, L.~Wang, and S.~Moriai, ``Privacy-preserving deep learning: Revisited and enhanced,'' in \emph{International Conference on Applications and Techniques in Information Security}.\hskip 1em plus 0.5em minus 0.4em\relax Springer, 2017, pp. 100--110.

\bibitem{fan2020rethinking}
L.~Fan, K.~W. Ng, C.~Ju, T.~Zhang, C.~Liu, C.~S. Chan, and Q.~Yang, ``Rethinking privacy preserving deep learning: How to evaluate and thwart privacy attacks,'' in \emph{Federated Learning}.\hskip 1em plus 0.5em minus 0.4em\relax Springer, 2020, pp. 32--50.

\bibitem{duff1989sparse}
I.~S. Duff, R.~G. Grimes, and J.~G. Lewis, ``Sparse matrix test problems,'' \emph{ACM Transactions on Mathematical Software (TOMS)}, vol.~15, no.~1, pp. 1--14, 1989.

\bibitem{paszke2019pytorch}
A.~Paszke, S.~Gross, F.~Massa, A.~Lerer, J.~Bradbury, G.~Chanan, T.~Killeen, Z.~Lin, N.~Gimelshein, L.~Antiga \emph{et~al.}, ``Pytorch: An imperative style, high-performance deep learning library,'' \emph{Advances in neural information processing systems}, vol.~32, 2019.

\bibitem{abadi2016tensorflow}
M.~Abadi, P.~Barham, J.~Chen, Z.~Chen, A.~Davis, J.~Dean, M.~Devin, S.~Ghemawat, G.~Irving, M.~Isard \emph{et~al.}, ``$\{$TensorFlow$\}$: a system for $\{$Large-Scale$\}$ machine learning,'' in \emph{12th USENIX symposium on operating systems design and implementation (OSDI 16)}, 2016, pp. 265--283.

\bibitem{krizhevsky2009learning}
A.~Krizhevsky, G.~Hinton \emph{et~al.}, ``Learning multiple layers of features from tiny images,'' Master's thesis, University of Toronto, 2009.

\bibitem{le2015tiny}
Y.~Le and X.~Yang, ``Tiny imagenet visual recognition challenge,'' \emph{CS 231N}, vol.~7, no.~7, p.~3, 2015.

\bibitem{lecun1998mnist}
Y.~LeCun, ``The mnist database of handwritten digits,'' \emph{http://yann. lecun. com/exdb/mnist/}, 1998.

\bibitem{he2016deep}
K.~He, X.~Zhang, S.~Ren, and J.~Sun, ``Deep residual learning for image recognition,'' in \emph{Proceedings of the IEEE conference on computer vision and pattern recognition}, 2016, pp. 770--778.

\bibitem{medmnistv2}
J.~Yang, R.~Shi, D.~Wei, Z.~Liu, L.~Zhao, B.~Ke, H.~Pfister, and B.~Ni, ``Medmnist v2-a large-scale lightweight benchmark for 2d and 3d biomedical image classification,'' \emph{Scientific Data}, vol.~10, no.~1, p.~41, 2023.

\bibitem{kairouz2021advances}
P.~Kairouz, H.~B. McMahan, B.~Avent, A.~Bellet, M.~Bennis, A.~N. Bhagoji, K.~Bonawitz, Z.~Charles, G.~Cormode, R.~Cummings \emph{et~al.}, ``Advances and open problems in federated learning,'' \emph{Foundations and Trends{\textregistered} in Machine Learning}, vol.~14, no. 1--2, pp. 1--210, 2021.

\bibitem{secagg_bell2020secure}
J.~H. Bell, K.~A. Bonawitz, A.~Gasc{\'o}n, T.~Lepoint, and M.~Raykova, ``Secure single-server aggregation with (poly) logarithmic overhead,'' in \emph{Proceedings of the 2020 ACM SIGSAC Conference on Computer and Communications Security}, 2020, pp. 1253--1269.

\bibitem{nguyen2022federated}
J.~Nguyen, K.~Malik, H.~Zhan, A.~Yousefpour, M.~Rabbat, M.~Malek, and D.~Huba, ``Federated learning with buffered asynchronous aggregation,'' in \emph{International Conference on Artificial Intelligence and Statistics}.\hskip 1em plus 0.5em minus 0.4em\relax PMLR, 2022, pp. 3581--3607.

\bibitem{diao2021heterofl}
E.~Diao, J.~Ding, and V.~Tarokh, ``Hetero{\{}fl{\}}: Computation and communication efficient federated learning for heterogeneous clients,'' in \emph{International Conference on Learning Representations}, 2021.

\bibitem{dwork2014algorithmic}
C.~Dwork, A.~Roth \emph{et~al.}, ``The algorithmic foundations of differential privacy,'' \emph{Foundations and Trends{\textregistered} in Theoretical Computer Science}, vol.~9, no. 3--4, pp. 211--407, 2014.

\bibitem{jayaraman2019evaluating}
B.~Jayaraman and D.~Evans, ``Evaluating differentially private machine learning in practice,'' in \emph{28th USENIX Security Symposium (USENIX Security 19)}, 2019, pp. 1895--1912.

\bibitem{wei2020federated}
K.~Wei, J.~Li, M.~Ding, C.~Ma, H.~H. Yang, F.~Farokhi, S.~Jin, T.~Q. Quek, and H.~V. Poor, ``Federated learning with differential privacy: Algorithms and performance analysis,'' \emph{IEEE Transactions on Information Forensics and Security}, vol.~15, pp. 3454--3469, 2020.

\bibitem{zhang2022understanding}
X.~Zhang, X.~Chen, M.~Hong, Z.~S. Wu, and J.~Yi, ``Understanding clipping for federated learning: Convergence and client-level differential privacy,'' in \emph{International Conference on Machine Learning, ICML 2022}, 2022.

\bibitem{secagg_kadhe2020fastsecagg}
S.~Kadhe, N.~Rajaraman, O.~O. Koyluoglu, and K.~Ramchandran, ``Fastsecagg: Scalable secure aggregation for privacy-preserving federated learning,'' \emph{arXiv preprint arXiv:2009.11248}, 2020.

\bibitem{zhao2021information}
Y.~Zhao and H.~Sun, ``Information theoretic secure aggregation with user dropouts,'' \emph{IEEE Transactions on Information Theory}, vol.~68, no.~11, pp. 7471--7484, 2022.

\bibitem{so2021turbo}
J.~So, B.~G{\"u}ler, and A.~S. Avestimehr, ``Turbo-aggregate: Breaking the quadratic aggregation barrier in secure federated learning,'' \emph{IEEE Journal on Selected Areas in Information Theory}, vol.~2, no.~1, pp. 479--489, 2021.

\bibitem{9712310}
A.~R. Elkordy and A.~S. Avestimehr, ``Heterosag: Secure aggregation with heterogeneous quantization in federated learning,'' \emph{IEEE Transactions on Communications}, vol.~70, no.~4, pp. 2372--2386, 2022.

\bibitem{wang2004image}
Z.~Wang, A.~C. Bovik, H.~R. Sheikh, and E.~P. Simoncelli, ``Image quality assessment: from error visibility to structural similarity,'' \emph{IEEE transactions on image processing}, vol.~13, no.~4, pp. 600--612, 2004.

\bibitem{russakovsky2015imagenet}
O.~Russakovsky, J.~Deng, H.~Su, J.~Krause, S.~Satheesh, S.~Ma, Z.~Huang, A.~Karpathy, A.~Khosla, M.~Bernstein \emph{et~al.}, ``Imagenet large scale visual recognition challenge,'' \emph{International journal of computer vision}, vol. 115, no.~3, pp. 211--252, 2015.

\end{thebibliography}

\appendices
\section{}
\label{sec:appendix}
\begin{table}[ht]
\vspace{-3mm}
\begin{center}
\begin{tabular}{|l|c|}
\hline
                       & \textbf{\begin{tabular}[c]{@{}c@{}}Leakage rate \\ (images) \end{tabular}} \\ \hline
\textbf{True initialization}     & 85.8\% (5492) \\
\textbf{Dataset agnostic} & 82.9\% (5305) \\
\textbf{Random}         & 62.1\% (3977) \\ \hline
\end{tabular}
\end{center}
\vspace*{-0mm}
\caption{\label{tab:bias_initialization} Leakage rate of \name using different bias initialization methods. 100 clients are trained on CIFAR-100 in FedAVG aggregation. The leakage rate only drops by $2.9\%$ for the dataset agnostic initialization.}
\vspace*{-5mm}
\end{table}

\begin{table}[ht]
\vspace{-3mm}
\begin{center}
\begin{tabular}{|l|cc|}
\hline
                       & \textbf{\begin{tabular}[c]{@{}c@{}}\name \\ FedSGD \end{tabular}} & \textbf{RtF FedSGD} \\ \hline
\textbf{CIFAR-100}     & 77.1\% (4936)                                                              & 77.1\% (4931)                                                             \\
\textbf{Tiny ImageNet} & 77.2\% (4939)                                                              & 77.7\% (4970) \\
\textbf{MNIST}         & 72.0\% (4610)                                                              & 75.1\% (4803)                                              
                                                            \\ \hline
\end{tabular}
\end{center}
\vspace*{-0mm}
\caption{\label{tab:fedsgd_leakage} Leakage rate of \name and Robbing the Fed (RtF)~\cite{fowl2022robbing} attack on several datasets in FedSGD. 100 clients in aggregation with a batch size of 64 are used.}
\vspace*{-8mm}
\end{table}

\subsection{Additional experiments}
\label{sec:appendix_additional}
\vspace{-2mm}
We show the leakage rate for \name under different bias initialization methods in Table~\ref{tab:bias_initialization}. 100 clients are trained on the CIFAR-100 dataset with 8 local iterations and local mini-batch size 8. FC layer size 256, $\textit{CSF}=500$, and $\alpha=1e-4$ are used. For average pixel intensity, the actual dataset distribution for CIFAR-100 is $\mu=0.4782$ and $\sigma=0.1470$. For the dataset agnostic initialization, we assume the server has no prior knowledge and the biases are initialized with $\mu=0.5$ and $\sigma=0.25$ as discussed in Section~\ref{sec:setting_biases}. The random method initializes all biases randomly between 0 and 1. The random initialization drops the leakage rate by $23.7\%$ while the dataset agnostic initialization only drops the overall leakage rate by $2.9\%$.

Table~\ref{tab:fedsgd_leakage} shows the leakage rate for \name and Robbing the Fed (RtF)~\cite{fowl2022robbing} in the FedSGD case on the CIFAR-100, Tiny ImageNet, and MNIST datasets. We use the same settings as in Section~\ref{sec:fedavg_leakage_rate} to allow for comparison of the leakage rate in FedAVG vs. FedSGD. For the FedSGD attacks, there are 100 clients in aggregation each with a batch size of 64. The \textit{CSF} value does not impact the FedSGD attacks, so we use $\textit{CSF}=1$. We use the same FC size/convolutional kernels as in the FedAVG attacks in Section~\ref{sec:fedavg_leakage_rate} for both methods. This also leads to a model size roughly $2\times$ larger for RtF in FedSGD. Despite this, both methods achieve very similar leakage rates. However, compared to the leakage rate of \name in FedAVG, both FedSGD attacks achieve a lower leakage rate on all datasets.

\begin{figure}[!t]
\begin{center}
\includegraphics[width=1.0\columnwidth,trim={0 0 0 10mm},clip]{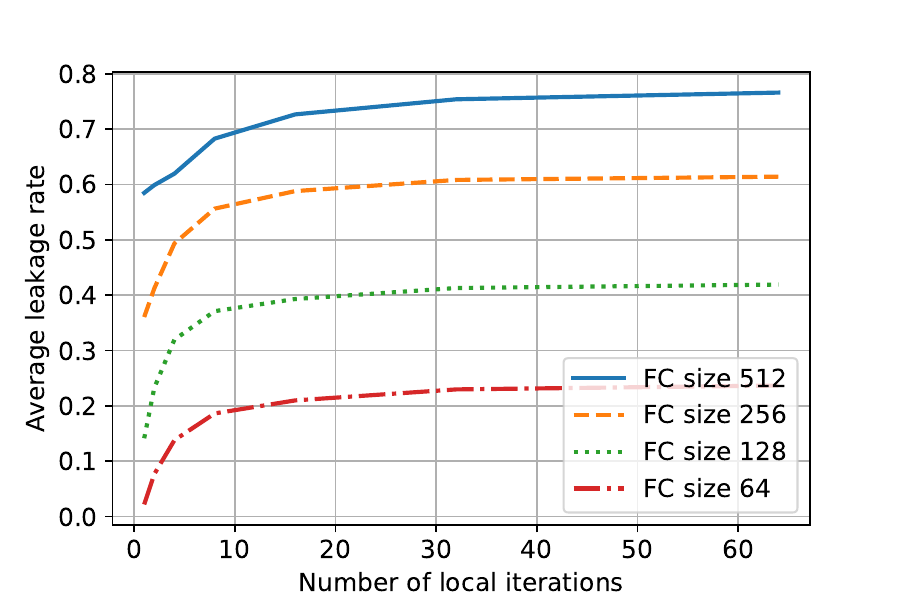}
\end{center}
\vspace*{-5mm}
\caption{\label{fig:dataset_size_local_iterations} Leakage rate based on the number of local iterations and FC layer size averaged over 10 clients. The local dataset size is fixed at 256 for CIFAR-100. Increasing the local iterations increases the leakage rate.}
\vspace*{-2mm}
\end{figure}

Figure~\ref{fig:dataset_size_local_iterations} shows the leakage rate for several FC layer sizes with a local dataset size of 256 when varying the number of local iterations averaged over 10 clients. Clients train on CIFAR-100 with $\textit{CSF}=500$ and $\alpha=1e-4$. Leakage rate is sampled with local iterations between 1-64, sampled by powers of 2. With a single local iteration, the attack leakage in FedAVG is the same as FedSGD with a batch size of 256. 

\vspace{-2mm}
\subsection{Secure aggregation in FL} \label{appendix-SA}
\vspace{-3mm}
Secure Aggregation \cite{bonawitz2017practical} is one of the core privacy-preserving techniques in FL, that 
enables the server to aggregate local model updates from a number of clients, without observing any of their individual model updates in the clear. At their core, state-of-the-art  secure aggregation protocols \cite{bonawitz2017practical,secagg_bell2020secure,secagg_so2021securing,secagg_kadhe2020fastsecagg,zhao2021information,so2021lightsecagg,so2021turbo,9712310} in FL rely on using  additive masking to protect the privacy of individual models. In particular, in a secure aggregation protocol, each user $i \in [N]$ encrypts  its own  model update   $\mathbf{y}^{(t)}_i = \text{Enc}(\mathbf{g}^{(t)}_i)$ before sending it to the server in  the $t$-th communication round.  This encryption is done such that secure aggregation guarantees the following:

 \noindent \textbf{Correct decoding.} The encryption  guarantees correct decoding  for the aggregated model such that the server should be able to decode 
\begin{equation}
    \text{Dec} \left(\sum_{i \in [N]} \mathbf{y}^{(t)}_i \right)= \sum_{i \in [N]} \mathbf{g}^{(t)}_i, 
\end{equation}
\noindent \textbf{Privacy guarantee.} The  encrypted model updates $\{\mathbf{y}^{(t)}_i\}
   _{i \in[N]}$ leak no information  about the model updates $\{\mathbf{g}^{(t)}_i\}
   _{i \in[N]}$ beyond the aggregated model $\sum_{i=1}^{N} \mathbf{g}^{(t)}_i$. This is formally given as the following 
   \begin{equation}\label{eq-SA_guarantee}
   I\left({\{\mathbf{y}}^{(t)}_i\}
   _{i \in[N]}; \{\mathbf{g}^{(t)}_i\}
   _{i \in[N]} \middle  |   \sum_{i=1}^{N} \mathbf{g}^{(t)}_i \right) = 0,
\end{equation}
where $I(.)$ represents the mutual information metric. 

\vspace{-2mm}
\subsection{Sample reconstructed images for other datasets}
\begin{figure}[t!]
\begin{center}
\includegraphics[width=1.0\columnwidth,trim={28mm 25mm 21mm 15mm},clip]{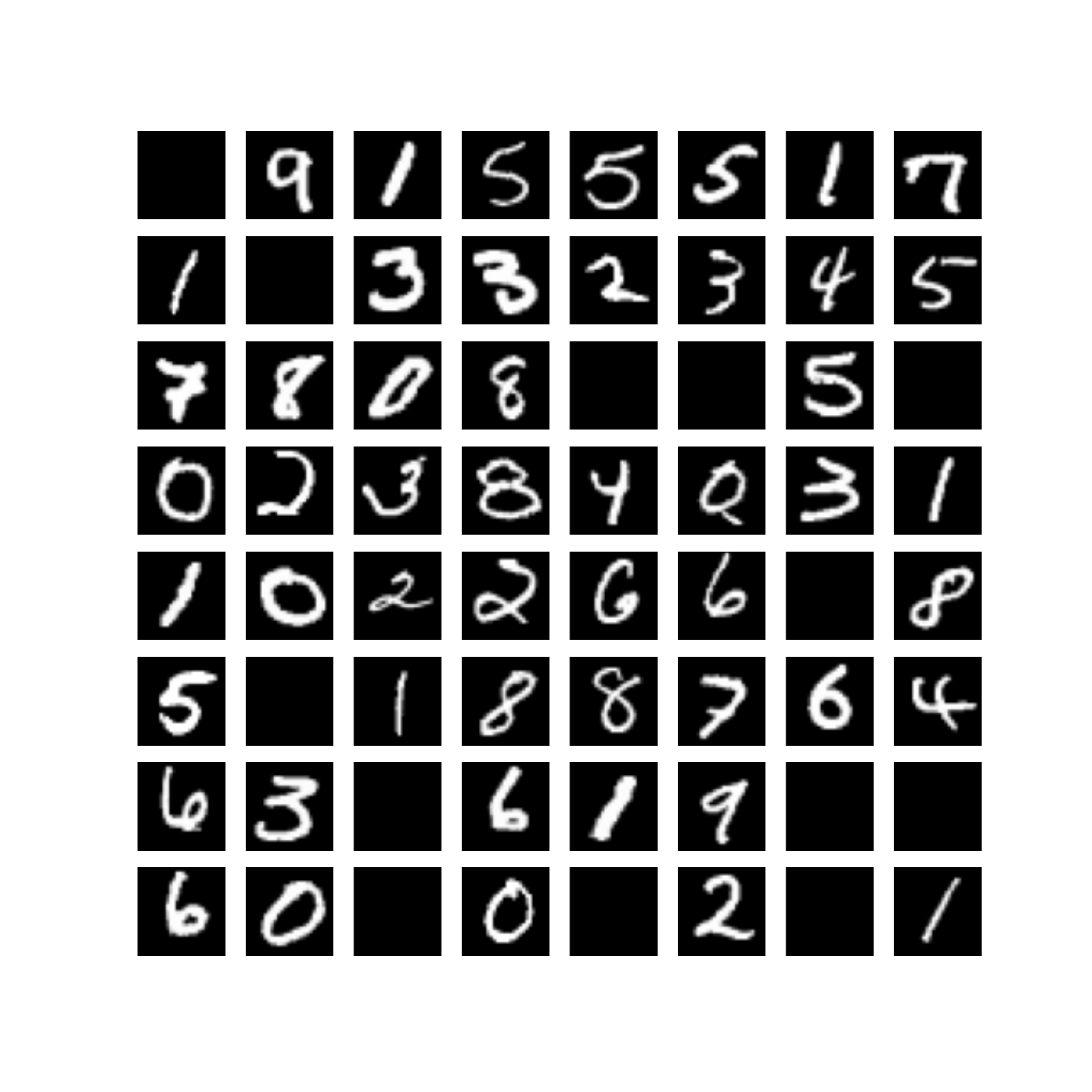}
\end{center}
\vspace*{-5mm}
\caption{\label{fig:single_client_mnist} Leaked images for a randomly chosen client training on the MNIST dataset. Out of the 64 total images, 51 images are leaked.}
\vspace{-5mm}
\end{figure}

\begin{figure}[t!]
\begin{center}
\includegraphics[width=1.0\columnwidth,trim={28mm 25mm 21mm 15mm},clip]{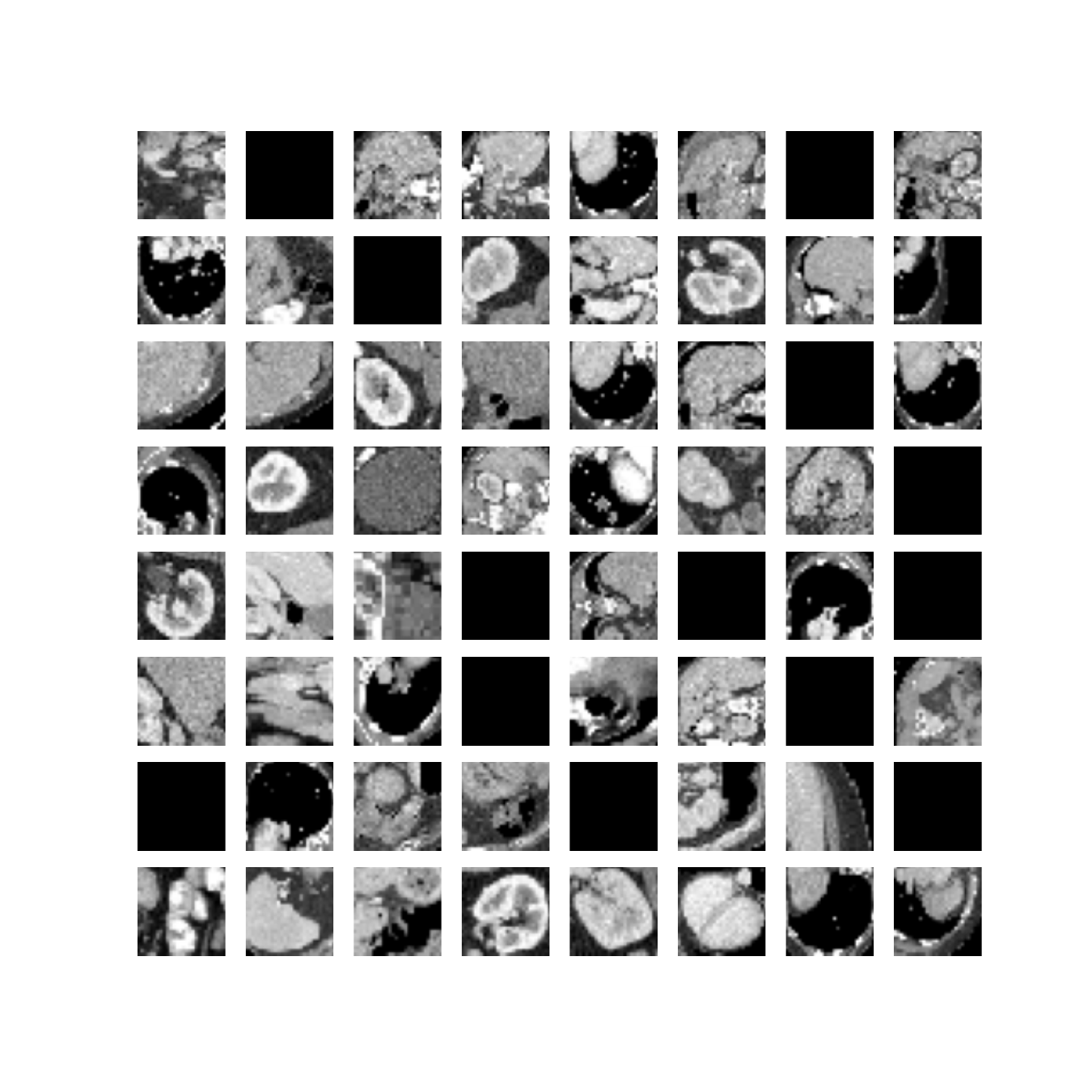}
\end{center}
\vspace*{-5mm}
\caption{\label{fig:single_client_organamnist} Leaked images for a randomly chosen client training on the OrganAMNIST dataset. Out of the 64 total images, 51 images are leaked.}
\vspace{-5mm}
\end{figure}

\begin{figure}[t!]
\begin{center}
\includegraphics[width=1.0\columnwidth,trim={28mm 25mm 21mm 15mm},clip]{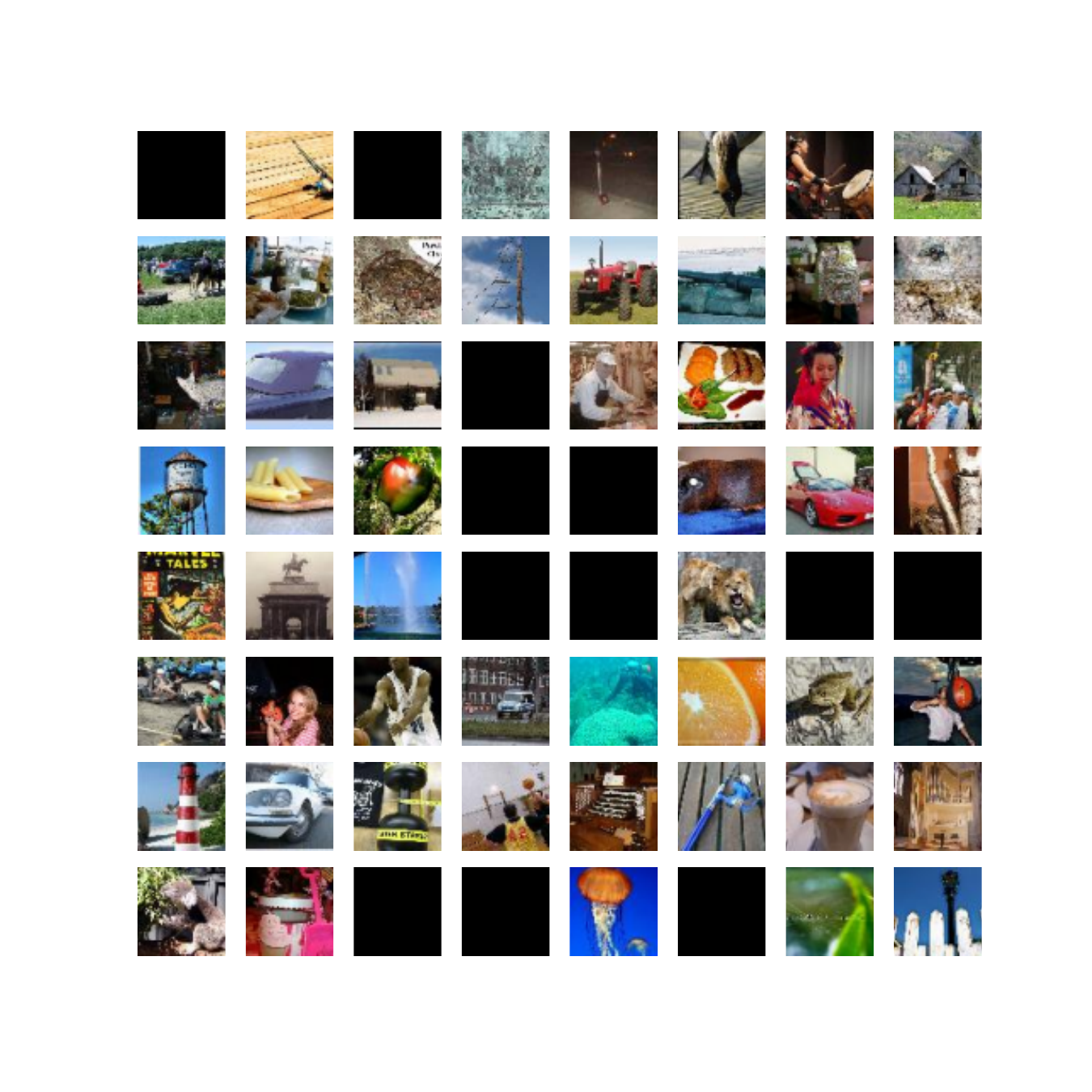}
\end{center}
\vspace*{-5mm}
\caption{\label{fig:single_client_tinyimagenet} Leaked images for a randomly chosen client training on the Tiny ImageNet dataset. Out of the 64 total images, 52 images are leaked.}
\vspace{-5mm}
\end{figure}

\begin{figure}[t!]
\begin{center}
\includegraphics[width=1.0\columnwidth,trim={28mm 25mm 21mm 15mm},clip]{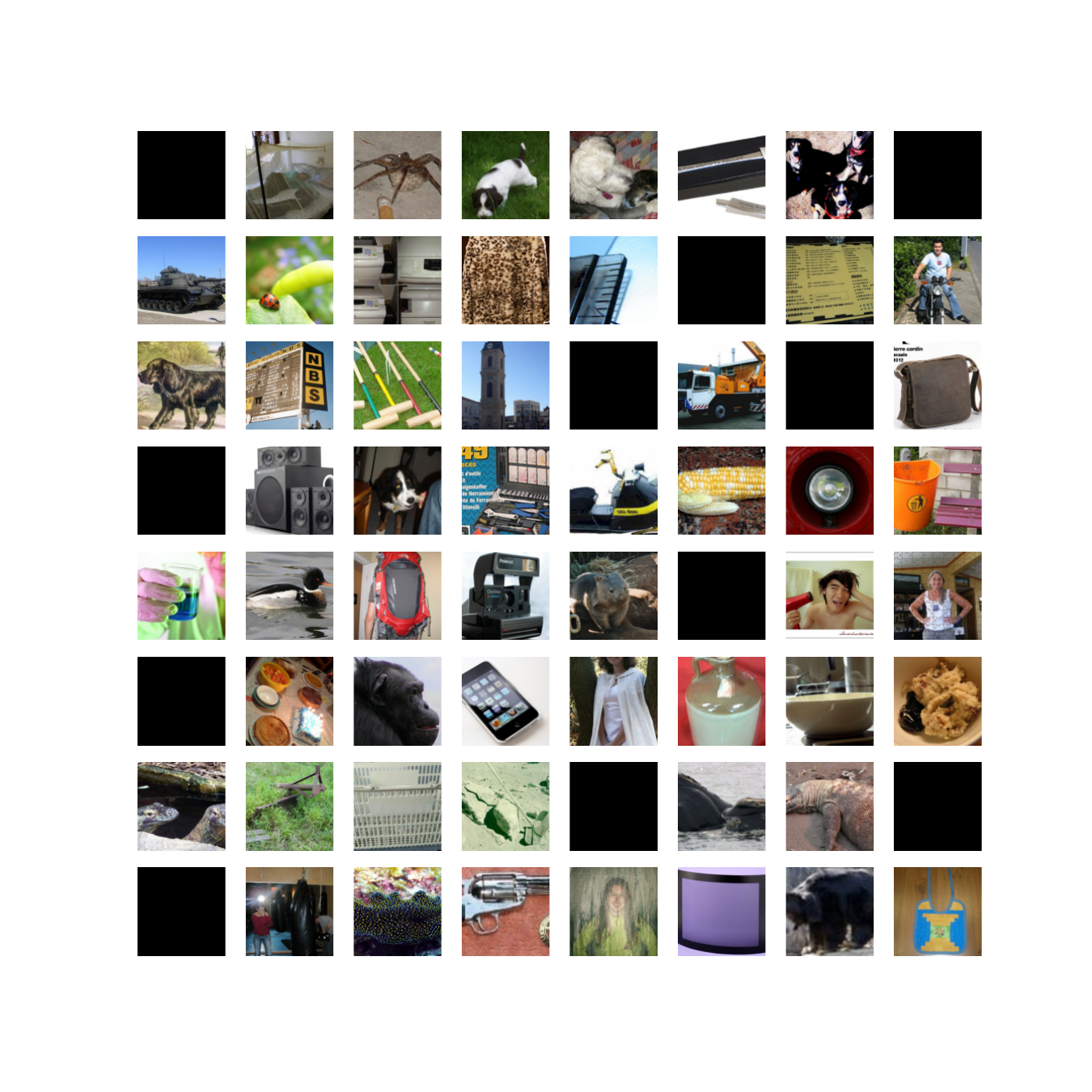}
\end{center}
\vspace*{-5mm}
\caption{\label{fig:single_client_imagenet} Leaked images for a single client training on the ImageNet~\cite{russakovsky2015imagenet} dataset. Out of the 64 total images, 53 images are leaked}
\end{figure}

\vspace{-2mm}
We show reconstructions of a randomly sampled client training on several datasets. There are 100 clients in aggregation training with 8 local iterations of mini-batch size 8, $\alpha=1e-4$, and $\textit{CSF}=100$. Figure~\ref{fig:single_client_mnist} shows the MNIST dataset, Figure~\ref{fig:single_client_organamnist} shows the OrganAMNIST dataset, Figure~\ref{fig:single_client_tinyimagenet} shows Tiny ImageNet, and Figure~\ref{fig:single_client_imagenet} shows ImageNet.

\begin{figure}[t!]
\vspace{-2mm}
\begin{center}     
\subfigure[Incorrect reconstruction]{\label{fig:psnr-ssim-1}\includegraphics[width=0.40\columnwidth]{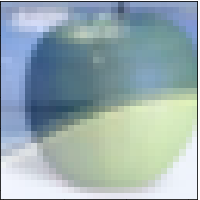}}\hspace{4mm}%
\subfigure[15.49 PSNR score]{\label{fig:psnr-ssim-2}\includegraphics[width=0.40\columnwidth]{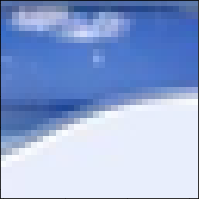}}\hspace{4mm}%
\subfigure[Correct reconstruction]{\label{fig:psnr-ssim-3}\includegraphics[width=0.40\columnwidth]{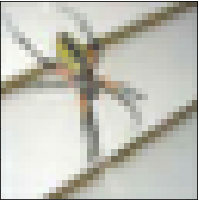}}\hspace{4mm}%
\subfigure[13.56 PSNR score]{\label{fig:psnr-ssim-4}\includegraphics[width=0.40\columnwidth]{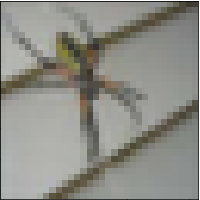}}
\end{center}
\vspace*{-3mm}
\caption{\label{fig:psnr-1} Reconstructed images from CIFAR-100 compared to the highest PSNR ground truth image. The incorrect reconstruction (a) has an overlap in the neuron activation while the correct reconstruction (c) has no overlap. The incorrectly reconstructed pair (a, b) has a higher PSNR score than the correct pair (c, d).}
\vspace*{-5mm}
\end{figure}

\vspace{-2mm}
\subsection{Image metrics}
\label{sec:image-metrics}
\vspace{-2mm}
For image reconstruction, the ability to identify an image and quantify the reconstruction quality are important for a metric to capture. We show from our attack results that the PSNR score achieves poor results in both categories. Previous works have also discussed that the use of mean squared error, which PSNR is based upon, is a poor reflection of image similarity. We use the SSIM~\cite{wang2004image} metric as a baseline for comparison and and show that it serves as a better metric for perceptual similarity. Our empirical results on the reconstructed images also support the idea that PSNR is not the best choice for a reconstruction quality metric, especially for linear layer leakage attacks. Particularly, when reconstructing an image involves a large shift in the pixel value range, PSNR functions very poorly. This section shows several cases to demonstrate this.

When observing the PSNR score of reconstructions, even if an image is reconstructed incorrectly, the PSNR score can be higher than the score for a correct reconstruction. Figure~\ref{fig:psnr-1} shows one such example, with two reconstruction results: the first being a failed reconstruction due to image overlap and the second being correct. Here the incorrectly reconstructed image has a higher PSNR score of 15.49 compared to the correct image which has a score of 13.56. The SSIM metric has the desired result, with a score of 0.91 for the correct image and 0.67 for the incorrect one.

\begin{figure}[t!]
\begin{center}     
\vspace{-2mm}
\subfigure[Reconstruction]{\label{fig:match-gt}\includegraphics[width=0.29\columnwidth]{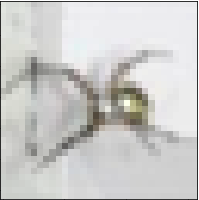}}\hspace{5mm}%
\subfigure[Max PSNR GT]{\label{fig:match-psnr}\includegraphics[width=0.29\columnwidth]{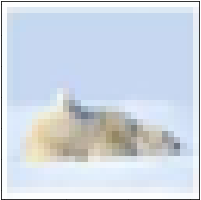}}\hspace{5mm}%
\subfigure[Max SSIM GT]{\label{fig:match-ssim}\includegraphics[width=0.29\columnwidth]{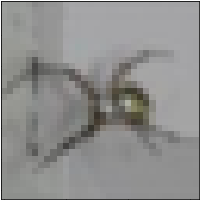}}
\end{center}
\vspace*{-3mm}
\caption{\label{fig:psnr-2} Matching a reconstructed image from CIFAR-100 to the ground truth image in the batch using the highest (b) PSNR  and (c) SSIM scores. SSIM chooses the correct image while PSNR chooses an incorrect image.}
\vspace*{-4mm}
\end{figure}

A more extreme case occurs if we use the highest PSNR score to match reconstructed images to their ground truth counterparts. Figure~\ref{fig:psnr-2} shows the result of choosing the corresponding ground truth image from the batch using the highest PSNR and SSIM scores. Comparing to the top ground truth image, the reconstruction in Figure~\ref{fig:match-psnr} has a PSNR score of 16.45 and an SSIM score of 0.17. The reconstruction in Figure~\ref{fig:match-ssim} has a PSNR score of 12.04 and a SSIM score of 0.88. As shown, using the maximum PSNR for image matching will often result in mismatches. In a batch of 100 images, there will usually be between 2-5 samples with an incorrect match. Comparatively, SSIM has no problems matching reconstructions to the ground truth.

\begin{figure}[ht]
\vspace{-2mm}
\begin{center}     
\subfigure[None]{\label{fig:downsample0}\includegraphics[width=0.29\columnwidth]{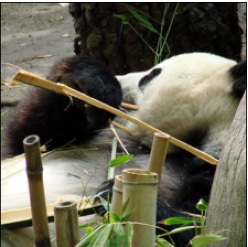}}\hspace{1.5mm}%
\subfigure[1$\times$]{\label{fig:downsample1}\includegraphics[width=0.29\columnwidth]{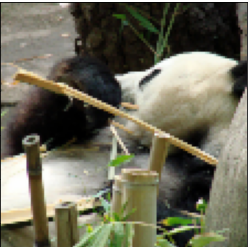}}\hspace{1.5mm}%
\subfigure[2$\times$]{\label{fig:downsample2}\includegraphics[width=0.29\columnwidth]{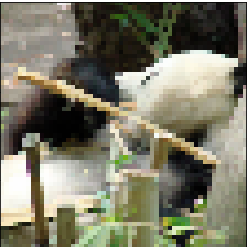}}%
\end{center}
\vspace*{-3mm}
\caption{\label{fig:imagenet-downsample} Sample from ImageNet dataset (a) without downsampling and after being downsampled (b) once and (c) twice. The number of max-pooling layers prior to reconstruction with an FC layer changes the amount of downsampling.}
\vspace*{-3mm}
\end{figure}

\begin{figure}[ht]
\vspace{-3mm}
\begin{center}     
\subfigure[OrganAMNIST]{\label{fig:distribution-organamnist}\includegraphics[width=0.47\columnwidth]{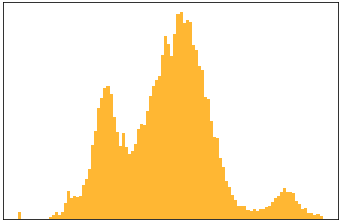}}\hspace{5mm}%
\subfigure[CIFAR-100]{\label{fig:distribution-cifar}\includegraphics[width=0.47\columnwidth]{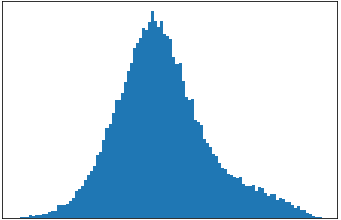}}\hspace{5mm}%
\end{center}
\vspace*{-3mm}
\caption{\label{fig:distribution-datasets} Average pixel intensity dataset distribution for (a) OrganAMNIST and (b) CIFAR-100.}
\vspace*{-3mm}
\end{figure}

\clearpage

\end{document}